\documentclass{article}
\PassOptionsToPackage{numbers, compress}{natbib}

\usepackage[preprint]{neurips_data_2024}

\usepackage[utf8]{inputenc} 
\usepackage[T1]{fontenc}    
\usepackage{hyperref}       
\usepackage{url}            
\usepackage{booktabs}       
\usepackage{amsfonts}       
\usepackage{nicefrac}       
\usepackage{microtype}      
\usepackage{colortbl}
\usepackage[table]{xcolor}
\usepackage{enumitem} 

\usepackage{microtype}
\usepackage{hyperref}
\usepackage{url}
\usepackage{bm}
\usepackage{booktabs}
\usepackage{pgfplots}
\pgfplotsset{compat=1.18}
\usepackage{pgfplotstable}
\usepackage{tikz}
\usetikzlibrary{patterns} 
\usepackage{verbatim}
\definecolor{darkblue}{rgb}{0, 0, 0.5}
\hypersetup{colorlinks=true, citecolor=darkblue, linkcolor=darkblue, urlcolor=darkblue}
\usepackage{graphicx}
\usepackage{multirow}
\usepackage{caption}
\usepackage{subcaption}
\usepackage{amsmath}
\newcommand{\Arrow}[1]{%
\parbox{#1}{\tikz{\draw[->](0,0)--(#1,0);}}
}
\usepackage{arydshln}

\newcommand*{\affaddr}[1]{#1}
\newcommand*{\affmark}[1][*]{\textsuperscript{#1}}
\newcommand*{\email}[1]{#1}

\definecolor{fred}{RGB}{255, 204, 204}
\definecolor{fpurple}{RGB}{204, 229, 255}
\definecolor{fyellow}{RGB}{204, 255, 204}
\definecolor{fgreen}{RGB}{229, 204, 255}
\definecolor{myorange}{RGB}{211, 84, 0}
\definecolor{flowpurple}{RGB}{150, 115, 166}
\definecolor{flowred}{RGB}{184, 84, 80}
\definecolor{flowyellow}{RGB}{214, 182, 86}
\definecolor{flowgreen}{RGB}{130, 179, 102}
\definecolor{lowlowblue}{RGB}{218, 232, 252}

\title{\textsc{SailCompass}: Towards Reproducible and Robust Evaluation for Southeast Asian Languages}

\author{Jia Guo\affmark[1]{\thanks{The first two authors contributed equally.}}~~,
    Longxu Dou\affmark[2]$^*$,
    Guangtao Zeng\affmark[3],
    Stanley Kok\affmark[1],
    Wei Lu\affmark[3],
    Qian Liu\affmark[2]\thanks{Corresponding author}\\
    \affaddr{\affmark[1]National University of Singapore};
    \affaddr{\affmark[2]Sea AI Lab};\\
    \affaddr{\affmark[3]Singapore University of Technology and Design}\\
    \email{
    \texttt{guojia@u.nus.edu, skok@comp.nus.edu.sg} \\
    \texttt{\{doulx, liuqian\}@sea.com}}\\
}

\begin{document}
\maketitle

\begin{abstract}

In this paper, we introduce SailCompass, a reproducible and robust evaluation benchmark for assessing Large Language Models (LLMs) on Southeast Asian Languages (SEA). 
SailCompass encompasses three main SEA languages, eight primary tasks including 14 datasets covering three task types (generation, multiple-choice questions, and classification). 
To improve the robustness of the evaluation approach, we explore different prompt configurations for multiple-choice questions and leverage calibrations to improve the faithfulness of classification tasks. 
With SailCompass, we derive the following findings:
(1) SEA-specialized LLMs still outperform general LLMs, although the gap has narrowed;
(2) A balanced language distribution is important for developing better SEA-specialized LLMs;
(3) Advanced prompting techniques (e.g., calibration, perplexity-based ranking) are necessary to better utilize LLMs.
All datasets and evaluation scripts are public\footnote{\url{https://github.com/sail-sg/sailcompass}}.

\end{abstract}

\section{Introduction}

Recent advancements in Large Language Models (LLMs) have led to numerous successful applications in language understanding and generation. 
However, current LLM research are mainly focus on English, Chinese and other Western languages, often overlooking other languages especially for low-resource languages like SEA languages.
Southeast Asia (SEA) is a vital region worldwide, comprising 11 countries and a population of approximately 675 million people (8.5\% of the world).

For NLP research, SEA region boasts remarkable linguistic diversity, which provides an ideal research ground for multilingual studies.
For instance, Indonesia alone has over 700 languages spoken daily~\cite{aji-etal-2022-one}, and also presents linguistic similarities, such as the shared terminology between Malay and Indonesian due to their intertwined history and culture.

Recently, SEA languages have witness several outstanding open models like SeaLLM~\cite{seallm2023}, Sealion~\cite{sealion2023} and Sailor~\cite{dou2024sailor}, which are built from scratch or continual pre-training on English-centric models.
However, most SEA benchmarks are either narrowed in task diversity or limited in examples scale, could not trustworthy measure the model performance.
We believe that a reproducible and robust evaluation system for LLMs, could largely assist the researchers to assess the the system's reliability and quantify the existing drawbacks.
It just likes a Compass for the Sailors to ship in the Sea.

In this paper, we present SailCompass, a reproducible and robust evaluation system for SEA languages in LLM era.
Our main contributions include benchmark, evaluation approache and evaluation findings.

\textbf{Comprehensive Evaluation Datasets}
(1) For evaluation faithfulness, we encompass three main SEA languages, eight primary tasks across three task formulations types (i.e., generation, multi-choice questions and classification), that demanding for both language proficiency and cultural understanding.
(2) For evaluation efficiency, we build SailCompass on the OpenCompass framework~\cite{2023opencompass}, for efficient running, extensible configuration and better visualization.

\textbf{Robust Evaluation Approach}
(1) For multiple-choice question tasks, we explore all prompt variants to identify the most robust configurations.
(2) For classification tasks, we leverage calibration to mitigate label bias and generate faithful outputs instead of meaningless random ones.

\textbf{Insightful Findings for Future Work}
(1)~We investigate the performance gap between general LLMs and SEA-specialized LLMs to analyse the future trend for application usage.
(2)~We highlight the importance of balanced language distribution for developing better SEA LLMs, certified by the performance of machine translation and text summarization.
(3)~We recognize the significance of advanced prompting technique (e.g., calibration, PPL-based ranking) for make better use of LLMs.

\section{Related Work}

This section focuses on SEA benchmarks. 
Refer to App.~\ref{appendix:multilingual_benchmarks} for general multilingual benchmarks.

\textbf{SEA benchmarks for Base Model}
NusaCrowd~\cite{NusaCrowd2023} introduced the first large Indonesian benchmark containing 137 datasets and over 200 tasks across 19 Indonesian languages. 
BHASA~\cite{BHASA2023} presented an evaluation suite containing linguistic and cultural evaluations for SEA languages. 
But BHASA only cares for zero-shot evaluation of API models like GPT-3.5, and only limited dataset size. 
SeaEval~\cite{SeaEval2024} additionally considers the cultural understanding ability of models, but the majority of the datasets are still based on high-resource languages, such as English and Chinese. 

\textbf{SEA benchmarks for Chat Model}
Sea-bench \citep{seallm2023} is a multilingual dataset with instructions across 9 SEA languages for evaluating chat model, examining five abilities like math, safety and task-solving. 
The linguists sourced the data by manually translating open-source English test sets, collecting real user questions from local forums and websites, collecting real math and reasoning questions, writing test instructions and questions themselves.

Distinguishing from these works, SailCompass aims to examine the instruction-following and few-shot learning ability of open-source LLMs through a considerable amount of tasks and datasets, with meticulous curation of datasets collected directly from native sources. 
Additionally, SailCompass places particular emphasis on evaluating tasks with multiple-choice targets and providing valuable insights through analysis.

In this work, SailCompass focus on open base model evaluation, covering three main SEA languages and 14 tasks, utilizing prompting and calibration to realize a robust evaluation benchmark.

\section{SailCompass Benchmark}

This section introduces \textsc{SailCompass}, an integrated evaluation suite, encompassing 
(1) the comprehensive multilingual benchmarks; 
(2) the reproducible evaluation codebase.

\subsection{Benchmark Construction Principle}

Prioritizing by the number of speakers, we consider three SEA languages: Indonesian, Vietnamese, and Thai. 
To build a comprehensive benchmark, we first identify four pivotal aspects for evaluating language models: 
(1) language proficiency; 
(2) reading comprehension; 
(3) reasoning ability; 
(4) cultural understanding. 
Based on these principles, we select a range of tasks and high-quality datasets.

In addition to task diversity, we prefer datasets created by native speakers and built on native corpora rather than translated from English benchmarks. 
These native benchmarks involve more localized entities, better examining the models' ability in cultural understanding and geographical knowledge. 
Unfortunately, such native-created datasets are limited, even after our best efforts to collect them. 
Thus, we also select high-quality multilingual datasets, like XNLI~\cite{xnli2018} and XQuAD~\cite{xquad2020}, to further supplement the number of datasets.

\begin{table}[t]
\centering
\small
\caption{Our benchmark includes eight tasks: Question Answering (QA), Machine Translation (MT), Text Summarization(TS), Examination (Exam), Commonsense Reasoning (CR), Machine Reading Comprehension (MRC), Natural Language Inference (NLI), and Sentiment Analysis (SA). 
}
\vspace{1mm}
\label{table:benchmark}
\begin{tabular}{ccccccc}
\toprule
\bf Type & \bf Datasets & \bf Thai & \bf Indonesian & \bf Vietnamese & \bf Task & \bf Domain \\
\midrule
\multirow{6}{*}{\textsc{GEN}} & \textsc{XQuAD}  & 1,149 & \multicolumn{1}{c}{--} & 1,169 &  & Wikipedia\\
& \textsc{TyDiQA} & \multicolumn{1}{c}{--} & 565 & \multicolumn{1}{c}{--} & \multirow{-2}{*}{QA} & Wikipedia \\
\arrayrulecolor{black!50}\cmidrule{2-7}
& \textsc{Flores-200} & 1,012 & 1,012 & 1,012 & MT & Wikipedia \\
\cmidrule{2-7}
& \textsc{ThaiSum} & 3,671 & \multicolumn{1}{c}{--} & \multicolumn{1}{c}{--} &  & News\\ 
& \textsc{IndoSum} & \multicolumn{1}{c}{--} & 3,762 & \multicolumn{1}{c}{--} &  & News\\
& \textsc{XLSUM} & \multicolumn{1}{c}{--} & \multicolumn{1}{c}{--} & 2,676 & \multirow{-3}{*}{TS} & News\\
\addlinespace[0.5ex]\arrayrulecolor{black}\midrule\addlinespace[0.5ex]
\multirow{3}{*}{\textsc{MCQ}} & \textsc{M3Exam} & 2,163 & 371~\footnote{Javanese} & 1,789 & Exam & School materials \\
& \textsc{XCOPA} & 500 & 500 & 500 & CR & General \\
& \textsc{BELEBELE} & 900 & 900 & 900 & MRC & General \\
\addlinespace[0.5ex]\midrule\addlinespace[0.5ex]
\multirow{5}{*}{\textsc{CLS}} & \textsc{XNLI} & 5,010 & \multicolumn{1}{c}{--} & 5,010 &  & Multi-genre\\
& \textsc{IndoNLI} & \multicolumn{1}{c}{--} & 5,182 & \multicolumn{1}{c}{--} & \multirow{-2}{*}{NLI} & Multi-genre\\
\arrayrulecolor{black!50}\cmidrule{2-7}
& \textsc{Wisesight} & 2,614 & \multicolumn{1}{c}{--} & \multicolumn{1}{c}{--} &  & Social media \\
& \textsc{Indolem} & \multicolumn{1}{c}{--} & 1,002 & \multicolumn{1}{c}{--} &  & Social media \\
& \textsc{VSMEC} & \multicolumn{1}{c}{--} & \multicolumn{1}{c}{--} & 692 & \multirow{-3}{*}{SA} & Social media \\
\arrayrulecolor{black}\bottomrule
\end{tabular}
\end{table}

\vspace{-1mm}
\subsection{Task and Dataset Construction}\label{sec:task_dataset_curation}
\vspace{-1mm}

We construct SailCompass on eight tasks across three task types.
\footnote{For datasets that no publicly available test set, we employ the validation set, like TydiQA.}
Refer to Table~\ref{table:benchmark} for statistics. 

\textbf{(1) Generation tasks (\textsc{GEN})}: Generate the token sequence given sequence input.
\vspace{-0.5em}
\begin{itemize}[leftmargin=2em, itemsep=0em]
\item \textbf{Question Answering (QA)} generates an answer span given the question and passage. 
We employ \textsc{XQuAD}~\cite{xquad2020} for Thai and Vietnamese and \textsc{TyDiQA} \citep{tydiqa2020} for Indonesian. 
\textsc{XQuAD} is translated from the English \textsc{SQuAD}~\cite{squad16}, while \textsc{TyDiQA} is built on native language corpus.

\item \textbf{Machine Translation (MT)} 

We employ \textsc{Flores-200}~\cite{flores2022}, which sourced data from web articles and annotated by professional translators.
We examine bi-directional translation performance between the target SEA languages and English, to examine the cross-lingual capability.

\item \textbf{Text Summarization (TS)} examines the ability to compress the key information from the paragraph. 
For Thai, we adopt \textsc{ThaiSum}\cite{thaisum2020}, and for Indonesian, we use \textsc{IndoSum}\cite{indosum2018}, both built on native language corpora. 
For Vietnamese, we adopt \textsc{XLSUM}~\cite{xlsum21} for evaluation.

\end{itemize}

\textbf{(2) Multiple-choice questions (\textsc{MCQ})}: Select the answer given a question and several options.

\vspace{-0.5em}
\begin{itemize}[leftmargin=2em, itemsep=0em]
\item  \textbf{Examination (Exam)} reflects the integrated intelligence of reasoning with domain knowledge. 
We use \textsc{M3Exam} \citep{m3exam2023}, built on school textbook, covering a range of subjects and levels.~\footnote{We use the Javanese split here, as M3Exam had not released the Indonesian split when submitting this paper.}

\item \textbf{Commonsense Reasoning (CR)} asks model to answer questions about commonsense understanding in daily scenario. 
We adopt \textsc{XCOPA} \citep{xcopa2020} for evaluation.

\item \textbf{Machine Reading Comprehension (MRC)} evaluates different levels of  language comprehension.
We adopt \textsc{BELEBELE} \citep{belebele2023} for evaluation, whose passage is from \textsc{Flores-200}~\cite{flores2022}.

\end{itemize}

\textbf{(3) Classification tasks (\textsc{CLS})}: Predict a label from a predefined set of categories.

\vspace{-0.5em}
\begin{itemize}[leftmargin=2em, itemsep=0em]
\item \textbf{Natural Language Inference (NLI)} 
Given a premise and a hypothesis, the model needs to select one correct answer from three labels to indicate their logical relation, i.e., \texttt{Entailment}, \texttt{Contradiction}, \texttt{Neutral}. 
We include \textsc{XNLI} \citep{xnli2018} for Thai and Vietnamese, and \textsc{IndoNLI} \citep{indonli2021} for Indonesian. 

\item \textbf{Sentiment Classification (SC)} aims to label the given text by different human feelings. 
We use \textsc{Wisesight}~\cite{wisesight2019} for Thai, \textsc{Indolem}~\cite{indolemsenti2020} for Indonesian, and \textsc{VSMEC}~\cite{vsmec} for Vietnamese.
They are all build the social media messages in native language.

\end{itemize}

\subsection{Instruction and Few-Shot Example Collection} 

For the selection of few-shot examples, we randomly select three examples from the training datasets or development datasets. 
For text summarization tasks, we only choose one example for demonstration.  
For each task, we employ a professional expert to write the task instructions in English and translate them by Google Translate or ChatGPT to the respective SEA languages. 

\subsection{Evaluation Protocol}
Our evaluation code are developed based on the open-source evaluation platform OpenCompass~\citep{2023opencompass}, an integrated framework that offers extensive configurations for assessing a broad range of large language models and datasets. 
It features efficient distributed evaluation to expedite processes and seamlessly incorporates support for new models and datasets.

We specialize in evaluating open-source base language models, which are easily accessible to a broad community. 
Our benchmarking efforts aim to enhance the transparency and reproducibility of large language model evaluations.
We employ greedy decoding for all experiments.

For Generation Tasks, we report BLEU~\cite{papineni-etal-2002-bleu} and Chrf++~\cite{popovic-2015-chrf}.
For MCQ Tasks, we report Exact Match.
For Classification Tasks, we report Exact Match and F1 Score.

\subsection{Evaluated Models}

All the evaluated models are base model, without instruction tuning and preference optimization (i.e., RLHF).
We don't consider the instruction/chat model, for estimating the upper bound of each model before post-training.

\footnote{We define a base model as one that is primarily trained on plain text, having weak instruction-following ability under a few-shot setting. SeaLLM did not release its base model. Thus, we adopt SeaLLM-Hybrid instead, which is obtained by training SeaLLM-base on a small set of instruction examples for security reasons.}
For model size, considering the trade-off between efficiency and effectiveness, we mainly consider the base model with size around 7B parameters. Refer to Table~\ref{table:model_link} for more details.

According to language distribution of the training corpus and model optimization methods, we categorized these models into three types:
\begin{itemize}
    \item  \textbf{General LLMs}: general multilingual models, whose training corpus cater for multilingual tokens, but mainly focus on western languages. It includes BLOOM~\cite{workshop2023bloom}, Llama-2~\cite{touvron2023llama},  Mistral~\cite{jiang2023mistral}, Qwen1.5~\cite{bai2023qwen}, Llama-3\footnote{\url{https://github.com/meta-llama/llama3}}, and Gemma~\cite{gemmateam2024gemma}.
    \item  \textbf{SEA-specific LLMs by continual pretraining}: train the General LLMs with SEA corpus, including VinaLLaMA~\cite{nguyen2023vinallama}, SeaLLM~\cite{seallm2023}, Sailor~\cite{dou2024sailor} and Typhoon~\cite{pipatanakul2023typhoon}.
    \item \textbf{SEA-specific LLMs by training from scratch}: training corpus consists of a significant number of SEA tokens and employ SEA friendly tokenizer, including Sea-Lion~\cite{sealion2023}.
\end{itemize}

For brevity, we collectively refer to the latter two groups of models as `SEA-specific LLMs'.

\section{Generation Tasks}
\label{sec:gen}

In this section, we focus on generation tasks, the most frequent reply format in real life. 
The evaluation for generation tasks is more robust than MCQ tasks and classification tasks. 
It's much easier for researchers to receive the consistent performance even under different configurations (e.g., altering the order of demonstrations, changing the number of demonstrations).
Thus, we suppose that generation tasks have been the mature tasks, serving as a satisfying measurement for base models.

We aims to explore the following aspects:
(1) the performance gap among general LLMs, SEA-specific LLMs and task-specific models, to see could we could solve the specific SEA tasks by developing LLMs;
(2) the pair-wise comparison between different model families (base model and cft model), to identify the key continual pre-training factors.

\begin{figure}
\centering
\begin{minipage}[b]{\textwidth}
\centering
\subcaptionbox{Machine Translation (En-X)\label{MT_EN_X}}{
\begin{tikzpicture}[scale=0.63]
\centering
\begin{axis}[
    ybar,
    width=10cm,
    height=4.2cm,
    bar width=0.3cm,
    ylabel style={yshift=1em},
    ylabel={Chrf++},
    symbolic x coords={
    \texttt{Qwen-1.5-7B}, 
    \texttt{Llama-3-8B}, 
    \texttt{Mistral-7B}, 
    \texttt{Gemma-7B}, 
    \texttt{Typhoon-8B}, 
    \texttt{NLLB-3.3B},
    \texttt{GPT3.5-turbo}, 
    \texttt{Sailor-7B}, 
    \texttt{SeaLLM-7B},
    \texttt{Sea-Lion-7B}, 
    },           
    xtick=data,
    nodes near coords align={vertical}, 
    legend style={at={(0.5, 1.2)}, anchor=north,legend columns=-1}, 
    ymin=0,
    x=2.1cm,
    enlarge x limits={abs=0.7cm},
    ]
    \addplot coordinates {
    (\texttt{Qwen-1.5-7B}, 57.25)
    (\texttt{Llama-3-8B}, 63.50)
    (\texttt{Mistral-7B}, 52.11)
    (\texttt{Gemma-7B}, 65.55)
    (\texttt{Typhoon-8B}, 58.42)
    (\texttt{NLLB-3.3B}, 68.8)
    (\texttt{GPT3.5-turbo}, 66.67)
    (\texttt{Sailor-7B}, 58.68)
    (\texttt{SeaLLM-7B}, 61.94)
    (\texttt{Sea-Lion-7B}, 58.68)
    };
    \addplot coordinates {
    (\texttt{Qwen-1.5-7B}, 29.88)
    (\texttt{Llama-3-8B}, 39.78)
    (\texttt{Mistral-7B}, 21.79)
    (\texttt{Gemma-7B}, 41.90)
    (\texttt{Typhoon-8B}, 42.98)
    (\texttt{NLLB-3.3B}, 40.5)
    (\texttt{GPT3.5-turbo}, 38.46)
    (\texttt{Sailor-7B}, 43.86)
    (\texttt{Sea-Lion-7B}, 36.57)
    (\texttt{SeaLLM-7B}, 36.94)
    };    
    \addplot coordinates {
    (\texttt{Qwen-1.5-7B}, 47.88)
    (\texttt{Llama-3-8B}, 56.30)
    (\texttt{Mistral-7B}, 38.93)
    (\texttt{Gemma-7B}, 56.97)
    (\texttt{Typhoon-8B}, 38.79)
    (\texttt{NLLB-3.3B}, 59.3)
    (\texttt{GPT3.5-turbo}, 55.81)
    (\texttt{Sailor-7B}, 58.11)
    (\texttt{Sea-Lion-7B}, 54.74)
    (\texttt{SeaLLM-7B}, 54.72)
    };   
    \legend{En\Arrow{.2cm}Id, En\Arrow{.2cm}Th, En\Arrow{.2cm}Vi} 
\end{axis}
\end{tikzpicture}
}
\strut
\end{minipage}
\begin{minipage}[b]{\textwidth}
\centering
\subcaptionbox{Machine Translation (X-En)\label{MT_X_EN}}{
\begin{tikzpicture}[scale=0.63]
\centering
\begin{axis}[
    ybar,
    width=10cm,
    height=4.2cm,
    bar width=0.3cm,
    ylabel style={yshift=1em},
    ylabel={BLEU},
    symbolic x coords={
    \texttt{Qwen-1.5-7B}, 
    \texttt{Llama-3-8B}, 
    \texttt{Mistral-7B}, 
    \texttt{Gemma-7B}, 
    \texttt{Typhoon-8B}, 
    \texttt{NLLB-3.3B},
    \texttt{GPT3.5-turbo}, 
    \texttt{Sailor-7B}, 
    \texttt{SeaLLM-7B},
    \texttt{Sea-Lion-7B}, 
    },           
    xtick=data,
    nodes near coords align={vertical}, 
    legend style={at={(0.5, 1.2)}, anchor=north,legend columns=-1}, 
    ymin=0,
    x=2.1cm,
    enlarge x limits={abs=0.7cm},
    ]

    \addplot coordinates {
    (\texttt{Qwen-1.5-7B}, 61.54)
    (\texttt{Llama-3-8B}, 64.76)
    (\texttt{Mistral-7B}, 61.45)
    (\texttt{Gemma-7B}, 66.17)
    (\texttt{Typhoon-8B}, 62.08)
    (\texttt{NLLB-3.3B}, 67.3)
    (\texttt{GPT3.5-turbo}, 38.0)
    (\texttt{Sailor-7B}, 64.58)
    (\texttt{SeaLLM-7B}, 62.68)
    (\texttt{Sea-Lion-7B}, 57.07)
    };    
    \addplot coordinates {
    (\texttt{Qwen-1.5-7B}, 49.70)
    (\texttt{Llama-3-8B}, 55.21)
    (\texttt{Mistral-7B}, 42.44)
    (\texttt{Gemma-7B}, 57.46)
    (\texttt{Typhoon-8B}, 55.44)
    (\texttt{NLLB-3.3B}, 56.8)
    (\texttt{GPT3.5-turbo}, 20.02)
    (\texttt{Sailor-7B}, 54.82)
    (\texttt{SeaLLM-7B}, 49.98)
    (\texttt{Sea-Lion-7B}, 46.63)
    };      
    \addplot coordinates {
    (\texttt{Qwen-1.5-7B}, 56.14)
    (\texttt{Llama-3-8B}, 58.20)
    (\texttt{Mistral-7B}, 51.22)
    (\texttt{Gemma-7B}, 60.00)
    (\texttt{Typhoon-8B}, 51.20)
    (\texttt{NLLB-3.3B}, 61.5)
    (\texttt{GPT3.5-turbo}, 28.84)
    (\texttt{Sailor-7B}, 58.64)
    (\texttt{SeaLLM-7B}, 55.01)
    (\texttt{Sea-Lion-7B}, 52.08)
    };      
    \legend{Id\Arrow{.2cm}En, Th\Arrow{.2cm}En, Vi\Arrow{.2cm}En} 
\end{axis}
\end{tikzpicture}
}
\strut
\end{minipage}
\caption{Machine translation results with Chrf++ as evaluation metric. 
}
\label{fig:results_machine_translation}
\end{figure}
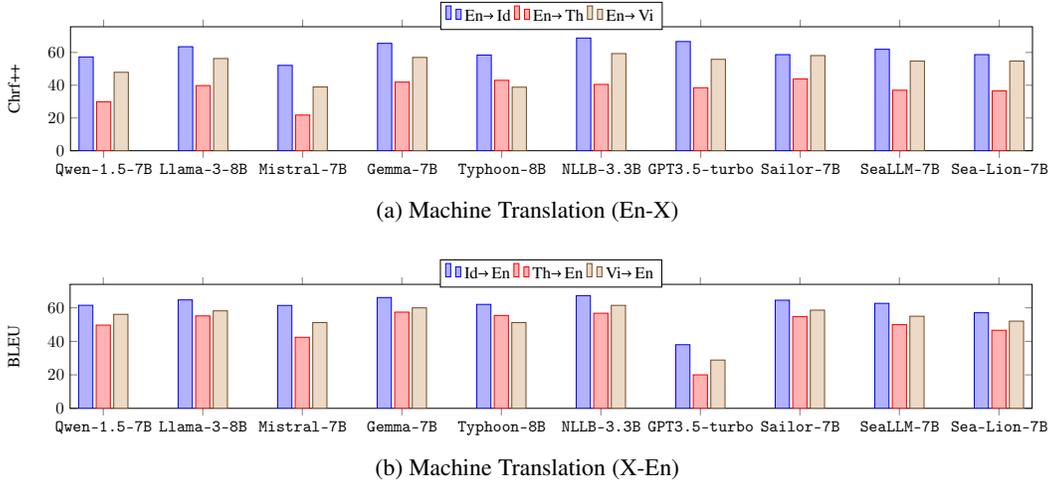

\subsection{Machine Translation}

We list the evaluation results with bi-direction in Figure~\ref{fig:results_machine_translation}.
We notice that Thai is really challenging in both directions for different models.
In the following, we summarize several useful findings.

\paragraph{English Prompt is Better Than Native Prompt}
(1) more balanced data distribution produces the more balanced model. Gemma and sailor, both have tiny difference between different language prompts;
(2) sealion pretraining corpus are mainly SEA languages, thus its native prompt is better than English prompt;
(3) overall, most model would benefit from the English prompt. 
We would recommend to use English prompt in the real machine translation scenarios.
But considering the usage scenario (SEA area prefer to use SEA languages rather than English), and the acceptable performance gap between native prompts and English prompts (less than 0.5 Chrf++), we will mainly report native prompts results in the following sections.

\paragraph{English-to-X is Harder Than X-to-English}
Generally, we suppose that `X->English' is much easier than `English->X', where X stands for a specific language and `->' indicates the the translation direction~\cite{Lu2023ChainofDictionaryPE}. 
With this criteria, we further explore the challenge of Thai.
First, as expected, `Thai->English' is always better than `English->Thai'. 
Then, we calculate the Chrf++ difference by `English->Thai' minus `Thai->English'.
We find this number is consistently larger than 10 across different model, which indicates the severe language degeneration happens in Thai.

\paragraph{Balanced Language Distribution Alleviates Language Degeneration}
We define a model has `language degeneration' problem, if its Chrf++ difference becomes larger after continual pre-training.
Based on this, we observe that the Chrf++ difference of sailor/seallm/sealion are obviously smaller that others.
It indicates that their balanced language distribution benefit the translation task.

\paragraph{LLMs Could Be Comparable with Specialized MT Models}
We compare the LLMs with the strong specialized baselines:
(1) NLLB-3.3B~\cite{team2022NoLL}, the specific machine translation model for more than 200 languages~\footnote{The results of NLLB-3.3B are from \url{https://tinyurl.com/nllb200dense3bmetrics} by NLLB Team.};
(2) GPT-3.5-Turbo, the well-known commerical LLM.\footnote{The results of GPT are from \citet{Lu2023ChainofDictionaryPE}.}
We observe that Llama/Gemma/Sailor are closed to basline performance, even they are built for general usage.

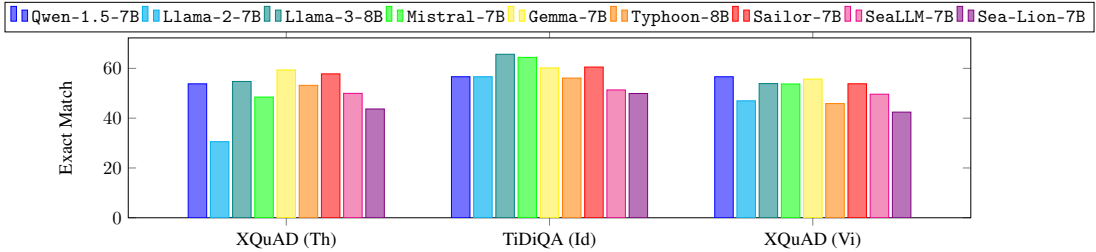
\begin{figure}
\begin{tikzpicture}[scale=0.7]
\centering
\begin{axis}[
    ybar,
    width=20cm,
    height=5cm,
    ylabel style={yshift=1em},
    ylabel={Exact Match},
    symbolic x coords={
    {XQuAD (Th)},
    {TiDiQA (Id)},
    {XQuAD (Vi)},
    },           
    xtick=data,
    nodes near coords align={vertical}, 
    legend style={at={(0.5, 1.2)}, anchor=north,legend columns=-1}, 
    ymin=0,
    x=5cm,
    enlarge x limits={abs=3cm},
    cycle list={%
        {blue,fill=blue!55},
        {cyan,fill=cyan!55},
        {teal,fill=teal!55},
        {green,fill=green!55},
        {yellow,fill=yellow!55},
        {orange,fill=orange!55},
        {red,fill=red!55},
        {magenta,fill=magenta!55},
        {violet,fill=violet!55},
    }
    ]
    \addplot 
    coordinates {
    ({XQuAD (Th)}, 53.79)
    ({TiDiQA (Id)}, 56.64)
    ({XQuAD (Vi)}, 56.63)
    };  
    \addplot 
    coordinates {
    ({XQuAD (Th)}, 30.55)
    ({TiDiQA (Id)}, 56.64)
    ({XQuAD (Vi)}, 46.96)
    };  
    \addplot 
    coordinates {
    ({XQuAD (Th)}, 54.74)
    ({TiDiQA (Id)}, 65.66)
    ({XQuAD (Vi)}, 53.89)
    };    	
    \addplot 
    coordinates {
    ({XQuAD (Th)}, 48.48)
    ({TiDiQA (Id)}, 64.42)
    ({XQuAD (Vi)}, 53.72)
    };      
    \addplot 
    coordinates {
    ({XQuAD (Th)}, 59.36)
    ({TiDiQA (Id)}, 60.18)
    ({XQuAD (Vi)}, 55.69)
    };    	
    \addplot 
    coordinates {
    ({XQuAD (Th)}, 53.18)
    ({TiDiQA (Id)}, 56.11)
    ({XQuAD (Vi)}, 45.85)
    };  	
    \addplot 
    coordinates {
    ({XQuAD (Th)}, 57.79)
    ({TiDiQA (Id)}, 60.53)
    ({XQuAD (Vi)}, 53.81)
    };  		
    \addplot 
    coordinates {
    ({XQuAD (Th)}, 49.96)
    ({TiDiQA (Id)}, 51.33)
    ({XQuAD (Vi)}, 49.62)
    }; 
    \addplot 
    coordinates {
    ({XQuAD (Th)}, 43.69)
    ({TiDiQA (Id)}, 49.91)
    ({XQuAD (Vi)}, 42.43)
    };  
    \legend{
    \texttt{Qwen-1.5-7B},
    \texttt{Llama-2-7B},
    \texttt{Llama-3-8B}, 
    \texttt{Mistral-7B}, 
    \texttt{Gemma-7B}, 
    \texttt{Typhoon-8B}, 
    \texttt{Sailor-7B}, 
    \texttt{SeaLLM-7B},
    \texttt{Sea-Lion-7B}, } 
\end{axis}
\end{tikzpicture}
\caption{Question Answering results with Exact Match as evaluation metric.
}
\label{fig:results_question_answering}
\end{figure}

\subsection{Question Answering}
The results are listed in Figure~\ref{fig:results_question_answering}. We observe that Llama-8b, Gemma-7b, Qwen.5-7b, and Sailor-7b demonstrate the best performances across all languages, according to their boost above average performance.

It's impressive that \textsc{TyDiQA} (Indonesian) receives the highest performance across all models, considering that TyDiQA is more challenging. \textsc{TyDiQA} is built directly from corpora in the target languages, taking into account more linguistic features and requiring the model to understand cultural and geographical knowledge, while \textsc{XQuAD} is translated from the English \textsc{SQuAD}~\cite{squad16}.
This conclusion aligns with Indonesian being a high-resource language (learning more common knowledge) and also sharing some word similarities with English (learning more linguistic knowledge).

\begin{table}[t]
\small
\begin{center}
\caption{
Main results on the text summarization task, measured in BLEU.
Each group contains the general LLM and their derived continual pre-training models.
As shown, continual pre-training generally boost the model performance in summarization, but it is challenging to achieve better performance on all languages.
}
\label{tab:text_summarization}
\scalebox{1.0}{
\begin{tabular}{crrr}
\toprule
\multirow{2}{*}{\bf Model}  & \multicolumn{3}{c}{\bf Text Summarization} \\
& \textsc{ThaiSum} (th) & \textsc{IndoSum} (id) & \textsc{XLSUM} (vi)  \\
\midrule

\texttt{Mistral-7B} & 16.78  &  46.55  &  6.02  \\
\hdashline
\texttt{Gemma-7B} & 14.29  &  36.65  &  4.70  \\
\hdashline
\texttt{BLOOM-7B1} & 0.24  &  26.76  &  2.26 \\
\hdashline
\texttt{Sea-Lion-7B} & 22.30  &  30.98  &  3.67 \\
\hdashline
\texttt{Qwen-1.5-7B} &  9.99  &  36.96  &  3.24   \\
\texttt{Sailor-7B} & 27.23  &  47.60  &  5.65 \\
\hdashline
\texttt{Llama-2-7B} &  17.10  &  48.34  &  4.67   \\
\texttt{VinaLLaMA-7B} & 2.60  &  39.24  &  5.92   \\
\texttt{SeaLLM-7B} & 18.50  &  48.68  &  4.60 \\
\hdashline
\texttt{Llama-3-8B} &  16.84  &  38.86  &  3.35 \\
\texttt{Typhoon-8B} & 19.37  &  38.66  &  2.44 \\
\bottomrule
\end{tabular}
}
\end{center}
\end{table}

\subsection{Text Summarization}
We suppose that the summarization task is one of the most challenging generation tasks, as it examines the ability to identify and compress key information from paragraph. 
This task could reflect the model's performance in complex real-world scenarios, such as long-context document question-answering and retrieval-augmented generation.

From Table~\ref{tab:text_summarization}, we observe that:
(1) Sailor-7B achieves the best overall performance, especially for Thai;
(2) Vietnamese is the most challenging language for summarization, with performance nearly half that of other languages;
(3) While the target language performances of Typhoon and Vinallama improve after continued pretraining, other languages degenerate. This indicates that their monolingual-specific continual pretraining greatly hurts the models' multilingual performance, highlighting the need for a balanced language distribution in the training data.

\section{Multiple-Choice Tasks}
\label{mcq}

Robust MCQ tasks evaluation faces three main challenges:
(1) \textbf{sensitivity to alterations}, including general prompt formatting~\cite{irrelevantContext2023, promptformatting2024} and MCQ-specific ones like option ordering~\cite{MCP2023};
(2) \textbf{option bias}~\cite{PriDe2024}, arising from intrinsic token bias, which becomes more severe when predicting only the option ID;
(3) \textbf{disparities across various evaluation criteria}, making it harder to align different evaluation results. 
For instance, the MCQ tasks evaluation mismatch rate is high between predictions derived from generated text output, and those based on first-token probability ranking~\citep{firsttoken2024}.

In this section, we analyze the key factors impacting the MCQ evaluation performance, then compare various evaluation criteria, and finally determine the best practice through extensive experiments.

\subsection{Investigation on Variants of MCQ Prompt}

\paragraph{Different Evaluation Approaches} 
We employ two approaches for evaluating MCQ: generative method (GEN-based) and the discriminative method (PPL-based). 
\textbf{GEN-based} inputs the prompt into the language model and takes the generation as prediction.
This approach aligns with realistic generation performance, but it might generate mismatched predictions.
\textbf{PPL-based} appends each option to the prompt, ranking the concatenated texts by perplexity scores, taking the lowest perplexity option as the answer. 
This approach avoids mismatches between generations and options.

\paragraph{Different Prompt Configurations} 
The key elements in composing MCQ prompt includes the options ID and options text, which appear in both the input and output.
These elements can be combined in different ways to generate various formats for the input (no-text\footnote{Equivalent to common question answering task under GEN-based evaluation approach.}, text-only, text with ID) and output (ID-only, text-only, text with ID). 
The combination of these formats composes different prompt configurations, which we refer to as different symbols for the 
input (e.g., ${T_i, L_iT_i}$) and output (e.g., ${L_o, T_o, L_oT_o}$).\
\footnote{Interpretion of symbols: T: option text, L: option label, i: model input, o: model output.}
Refer to Appendix~\ref{promp_variants} for detailed explanation of five prompt variants.

\paragraph{Comparison with Different Configurations under Different Approaches}
Under various configurations, we evaluate General LLMs and SEA-specific LLMs on Thai split of BELEBELE (Table~\ref{prompt_main}).
We notice significant variations in model performance when the prompt changes: 
(1) when LLMs are required to give the answer text as the prediction (${T_o}$), introducing either option content (${L_iT_iT_o}$) or the option ID (${T_iT_o}$) in the prompt cause performance degradation. 
(2) For ${L_iT_iL_o}$, Gen-based eval and PPL-based eval are nearly identical across various models, which illustrates that the choice to adopt either generative or discriminative methods is not the primary factor causing the performance fluctuations. 
(3) Compared to the setting ${T_o}_{_{PPL}}$, it is worth noting that the results of \verb|Mistral-7B| and \verb|Sailor-7B| are significantly improved by 40\% and 47 \%, respectively, when models are restricted to predicting the correct option ID in the setting of PPL-based (${L_iT_iL_o}_{_{PPL}}$), evaluating \verb|Mistral-7B| to the second best position for Southeast Asian languages.

\begin{table}[ht]
\centering
\caption{Experiments on the BELEBELE (th) benchmark measured in Exact Match (EM).}
\label{prompt_main}
\begin{tabular}{lcccccc}
\toprule
\multicolumn{1}{c}{\bf Model}  &\multicolumn{1}{c}{\tiny ${\bm{T_o}}_{_{PPL}}$} &\multicolumn{1}{c}{\tiny ${\bm{T_iT_o}}_{_{PPL}}$} &\multicolumn{1}{c}{\tiny ${\bm{L_iT_iT_o}}_{_{PPL}}$} &\multicolumn{1}{c}{ \tiny ${\bm{L_iT_iL_o}}_{_{PPL}}$} &\multicolumn{1}{c}{\tiny ${\bm{L_iT_iL_o}}_{_{GEN}}$}  &\multicolumn{1}{c}{\tiny ${\bm{L_iT_iL_oT_o}}_{_{PPL}}$} \\
\cmidrule[0.5pt](lr){1-1} \cmidrule[0.5pt](lr){2-4} \cmidrule[0.5pt](lr){5-6} \cmidrule[0.5pt](lr){7-7}
\verb|Llama-2-7B|  & 32.44 & 25.22 & 26.89 & 26.11 & 27.56 & 25.11\\
\verb|Mistral-7B|  & 34.44 & 28.11 & 30.33 & 48.22 & 49.00 & 28.11\\
\verb|Sea-Lion-7B|  & 36.78 & 25.78 & 27.44 & 25.33 & 25.44 & 25.56\\
\verb|SeaLLM-7B|   & 37.44 & 28.56 & 29.56 & 38.22 & 38.44 & 28.33 \\
\verb|Sailor-7B|   & 42.56 & 32.44 & 36.44 & 62.56 & 62.78 & 33.89 \\
\bottomrule
\end{tabular}
\end{table}

\subsection{Measure the Robustness of MCQ Prompt Varients}
We first propose three measures in evaluating the robustness of prompt variants: 
(1) mitigating token bias towards specific option IDs;  
(2) remaining unbiased regarding the surface form of options; 
(3) resisting manipulation intended to inflate model performance.

\begin{figure}
\centering
\begin{minipage}[b]{0.45\textwidth}
\centering
{\begin{tikzpicture}[scale=0.75]
\centering
\begin{axis}[
    ybar,
    bar width=0.25cm,
    ylabel style={yshift=-0.2em},
    ylabel={Recall std},
    symbolic x coords={
        ${T_o}$, 
        ${T_iT_o}$,
        ${L_iT_iT_o}$,
        ${L_iT_iL_o}$,
        ${L_iT_iL_oT_o}$,   
    },
    xtick=data,
    nodes near coords align={vertical}, 
    legend style={at={(0.5, 1)}, anchor=north,legend columns=-1}, 
    ymin=0,
    ymax=0.3
    ]
    \addplot coordinates {
        (${T_o}$, 0.0169) 
        (${T_iT_o}$, 0.0422) 
        (${L_iT_iT_o}$, 0.0307) 
        (${L_iT_iL_o}$, 0.2427) 
        (${L_iT_iL_oT_o}$, 0.0515)
    };  
    \addplot coordinates {
        (${T_o}$, 0.0531) 
        (${T_iT_o}$, 0.0221) 
        (${L_iT_iT_o}$, 0.0144) 
        (${L_iT_iL_o}$, 0.166) 
        (${L_iT_iL_oT_o}$, 0.0172)
    };  
    \addplot [pattern=north east lines, pattern color=blue] coordinates {
        (${T_o}$, 0.0186) 
        (${T_iT_o}$, 0.0219) 
        (${L_iT_iT_o}$, 0.0319) 
        (${L_iT_iL_o}$, 0.0995) 
        (${L_iT_iL_oT_o}$, 0.0426)
    };  
    \legend{\texttt{Sea-Lion-7B}, \texttt{SeaLLM-7B}, \texttt{Sailor-7B}}               
\end{axis}
\end{tikzpicture}}
\strut
\end{minipage}
\vspace{-0.5cm}
\begin{minipage}[b]{0.45\textwidth}
\centering
{

\begin{tikzpicture}[scale=0.75]
\centering
\begin{axis}[
    ybar,                                           
    bar width=0.25cm,
    ylabel style={yshift=-0.2em},
    ylabel={Relative length discrepancy (\%)},                                
    symbolic x coords={
    ${T_o}$, 
    ${T_iT_o}$,
    ${L_iT_iT_o}$,
    ${L_iT_iL_o}$,
    ${L_iT_iL_oT_o}$, 
    },                  
    xtick=data,                                     
    nodes near coords align={vertical},             
    legend style={at={(0.5, 1)}, anchor=north,legend columns=-1}, 
    ymin=0,
    ymax=45
    ]
    \addplot coordinates {
    (${T_o}$, 7.43) 
    (${T_iT_o}$, 26.72) 
    (${L_iT_iT_o}$, 26.86) 
    (${L_iT_iL_o}$, 0.44) 
    (${L_iT_iL_oT_o}$, 26.24)};  
    \addplot coordinates {
    (${T_o}$, 12.13) 
    (${T_iT_o}$, 30.5) 
    (${L_iT_iT_o}$, 30.24) 
    (${L_iT_iL_o}$, 4.1) 
    (${L_iT_iL_oT_o}$, 31.06)};  
    \addplot [pattern=north east lines, pattern color=blue] coordinates {
    (${T_o}$, 13.99) 
    (${T_iT_o}$, 32.28) 
    (${L_iT_iT_o}$, 31.78)
    (${L_iT_iL_o}$, 1.73) 
    (${L_iT_iL_oT_o}$, 32.3)};  
    \legend{\texttt{Sea-Lion-7B}, \texttt{SeaLLM-7B}, \texttt{Sailor-7B}}                
\end{axis}
\end{tikzpicture}}
\strut
\end{minipage}
\caption{
Analysis of prediction bias across prompt variants, with PPL-based evaluation approach.
}
\label{fig:prediction_bias}
\end{figure}
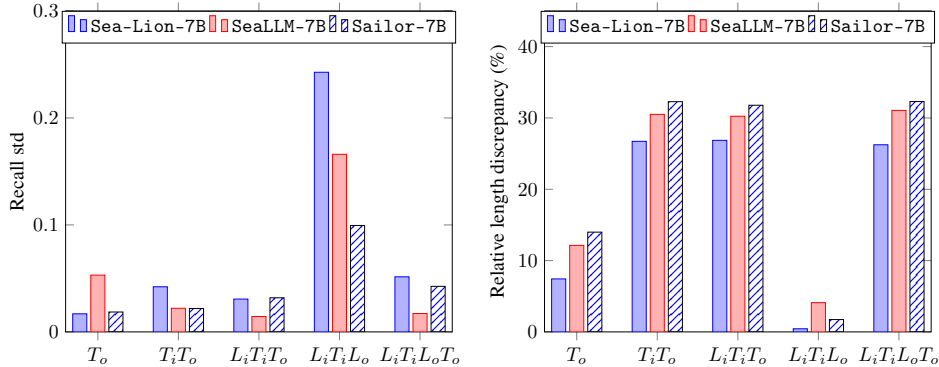

\paragraph{Does MCQ evaluation favor specific option IDs?} 
Following \citet{PriDe2024}, we adopt the standard deviation of recall across options as the metric to measure option ID bias
As shown in Figure~\ref{fig:prediction_bias} (left), ${L_iT_iL_o}$ exhibits the highest standard deviation in recall. 
This suggests that ${L_iT_iL_o}$ influences the model to produce imbalanced predictions towards specific option IDs. 
In contrast, the prompts that include the option text (i.e., containing ${T_o}$) mitigate the potential biases associated with option IDs. 
Additionally, different models demonstrate distinct preferences for different prompts. For instance, \verb|SEA-LION-7B| and \verb|Sailor-7B| exhibit low performance, while \verb|SeaLLM-7B| performs well.

\paragraph{Does MCQ evaluation prefer specific option surface forms?} 
We use character-level length to measure surface form bias.  Figure~\ref{fig:prediction_bias} (right) shows the relative length discrepancy (\%) of model predictions with prompt variants, indicating whether prediction lengths overestimate or underestimate reference options. Incorporating option text like ${T_o}$ biases models towards longer options. ${T_o}$ exhibits lower bias, indicating greater resilience against length biases. ${L_iT_iL_o}$, only outputting option ID, is unaffected by length bias but introduces severe token bias as discussed above.

\begin{table}[t]
\small
\caption{
Experiments on MCQ benchmark measured in Exact Match (EM) with manipulated training.
}
\label{qwen_mmlu}
\begin{center}
\begin{tabular}{lrrrrrrrrr}
\toprule
\multirow{3}{*}{\bf M3Exam}
& \multicolumn{3}{c}{${\bm{T_o}}_{_{PPL}}$} & \multicolumn{3}{c}{${\bm{L_iT_iL_o}}_{_{PPL}}$} & \multicolumn{3}{c}{${\bm{L_iT_iL_o}}_{_{GEN}}$}
\\
\cmidrule(lr){2-4} \cmidrule(lr){5-7}  \cmidrule(lr){8-10}
& th & jv & vi & th & jv & vi & th & jv & vi \\
\midrule
\textsc{Qwen1.5-7B}  & 25.75 & 26.15 & 36.28 & 36.89 & 31.54 & 51.31 & 35.88 & 32.35 & 51.09 \\
\textsc{Qwen1.5-7B$_\mathcal{M}$} & 27.23 & 26.15 & 36.61 & 37.49 & 34.50 & 52.93 & 36.80 & 32.61 & 52.88 \\
\midrule
\multicolumn{1}{c}{$\triangle_{avg}$} & \multicolumn{3}{c}{0.07} & \multicolumn{3}{c}{0.14} & \multicolumn{3}{c}{0.07} \\
\bottomrule
\end{tabular}

\vspace{2mm}

\begin{tabular}{lrrrrrrrrr}
\toprule
\multirow{3}{*}{\bf BELEBELE} 
& \multicolumn{3}{c}{${\bm{T_o}}_{_{PPL}}$} & \multicolumn{3}{c}{${\bm{L_iT_iL_o}}_{_{PPL}}$} & \multicolumn{3}{c}{${\bm{L_iT_iL_o}}_{_{GEN}}$}
\\
\cmidrule(lr){2-4} \cmidrule(lr){5-7}  \cmidrule(lr){8-10} 
& id & vi & th & id & vi & th & id & vi \\
\midrule
\textsc{Qwen1.5-7B}  & 37.89 & 42.11 & 42.56 & 64.33 & 72.11 & 73.78 & 47.89 & 62.22 & 44.44\\
\textsc{Qwen1.5-7B$_\mathcal{M}$} & 37.89 & 40.67 & 42.33 & 64.78 & 72.89 & 77.00 & 58.00 & 71.11 & 71.33\\
\midrule
\multicolumn{1}{c}{$\triangle_{avg}$} & \multicolumn{3}{c}{-0.04} & \multicolumn{3}{c}{0.06} & \multicolumn{3}{c}{0.96} \\
\bottomrule
\end{tabular}
\end{center}
\end{table}

\paragraph{Can evaluation performance be manipulated?}

From the above analysis, we prioritize two prompt configurations: ${T_o}$ and ${L_iT_iT_o}$. 
${T_o}$ demonstrates resilience to selection biases in both option ID and length. 
${L_iT_iT_o}$ is robust to option length bias but more sensitive to option ID bias. 
To determine the final configuration, we conduct a manipulated training experiment using a subset of MMLU~\cite{mmlu2021} (MCQ datasets) to fit the evaluation prompt format (Table~\ref{qwen_mmlu}). 
Refer to Appendix~\ref{appendix:manipulate_exp} for details. 
With manipulated training, the model shows a significant 17.2\% improvement with prompt ${L_iT_iL_o}$. 
Interestingly, SEA languages performance improves with manipulated training on English data but weakens the reliability of ${L_iT_iL_o}$. 
In contrast, ${T_o}$ performance decreases by about 1.8\% with manipulated training, suggesting challenges in achieving higher performance with scaled-up training.

In summary, we employ the prompt configuration ${\bm{T_o}}_{_{PPL}}$ as the recommended evaluation approach for MCQ tasks of SailCompass, to ensure a robust evaluation.

\section{Classification Tasks}
\label{sec:cls}

\begin{table}[t]
\begin{center}
\caption{Main results on the classification task measured in Exact Match (EM).
}
\label{cls_task}
\scalebox{0.8}{
\begin{tabular}{crrrrrr}
\toprule
\multirow{2}{*}{\bf Model}  & \multicolumn{3}{c}{\bf Natural Language Inference} & \multicolumn{3}{c}{\bf Sentiment Classification} \\
\cmidrule(lr){2-4} \cmidrule(lr){5-7}
& \textsc{XNLI} (th) & \textsc{IndoNLI} (id) & \textsc{XNLI} (vi) & \textsc{Wisesenti} (th) & \textsc{Indolem} (id) & \textsc{VSMEC} (vi) \\
\midrule
\rowcolor{gray!30}
Random &  33.33 &  33.33 &  33.33 &  33.33 &  50.00 &  14.29 \\
\texttt{Llama-3-8B} &  35.25  &  35.31 &  36.19  &   46.44  &  86.03 &  19.51 \\
\texttt{Mistral-7B} & 31.96  &  33.62 &  33.67    &   46.25  &  70.86 &  9.1 \\
\texttt{Gemma-7B} & 35.63  &  36.92 &  35.71   &   43.23  &  85.23 &  20.81 \\
\texttt{Falcon-7B} & 33.75  &  33.44 &  34.19   &  48.74  &  79.84 &  6.65   \\
\texttt{Qwen-1.5-7B} &  36.47  &  35.30 &  38.28  & 35.73  &  83.03 &  21.24   \\
\texttt{Llama-2-7B} &  32.65  &  33.11 &  33.35  & 51.87  &  81.54 &  18.5   \\
\texttt{Typhoon-8B} & 34.85  &  34.20 &  33.49  &   30.64  &  72.75 &  7.51\\
\texttt{VinaLLaMA-7B} & 30.94  &  35.14 &  37.74  &   47.05   &  30.94 &  20.09 \\
\texttt{Sailor-7B} & 35.89  &  37.88 &  34.89   &   32.44  &  80.44 &  24.42 \\
\texttt{SeaLLM-7B} & 35.77  &  35.22 &  34.27  &   31.71  &  76.15 &  20.23 \\
\texttt{Sea-Lion-7B} & 33.65  &  34.74 &  32.50   &   32.36  &  84.73 &  18.35 \\
\bottomrule
\end{tabular}
}
\end{center}
\end{table}

Compared with generation tasks and MCQ tasks, classification tasks are the most challenging ones. 
They are more sensitive to evaluation factors due to majority label bias, common token bias, and recency bias~\cite{Zhao2021CalibrateBU}. 
This inevitably prevents researchers from obtaining trustworthy evaluation results.

In this section, we address the challenge with \textbf{Contextual Calibration} \cite{Zhao2021CalibrateBU}, which can effectively increase the probability of generating convincing predictions, thus avoiding the underestimation issue.
First, we estimate the bias via context-free test input, then counter the bias by normalizing the SoftMax scores of label. Thus alleviate the underestimation problem by label bias.
We present more detail about the calibration in Appendix~\ref{appendix:calibration}.

\paragraph{Results}
The results for NLI and sentiment classification tasks are shown in Table~\ref{cls_task}.
We notice that base models can not achieve a good performance in NLI tasks with marginally improvement compared to random choose.
Compared to NLI task, language models can achieve better results on sentiment classification tasks.

\section{Conclusion}
In this work, we present SailCompass, a comprehensive suite of evaluation scripts designed for robust and reproducible evaluation of multilingual language models targeting Southeast Asian languages. SailCompass encompasses three major SEA languages and covers eight primary tasks using 14 datasets, spanning three task types: generation, multiple-choice questions, and classification.
To enhance the robustness of our evaluation approach, we explore different prompt configurations for multiple-choice questions and employ calibration methods to improve the accuracy of classification tasks. 
Through SailCompass, we have derived several key findings regarding model continual pretraining and robust model evaluation.

We believe that SailCompass will be highly beneficial for the development of large language models tailored to the Southeast Asia region, providing a crucial resource for researchers in this area.

\section*{Limitations}

For the evaluation benchmark:
\begin{enumerate}
    \item \textbf{Model Type}: The current version of SailCompass focuses solely on base models for few-shot evaluation. Future versions should also consider incorporating chat tasks for zero-shot evaluation.
    \item \textbf{Language Coverage}: SailCompass currently supports three languages. Future work should expand coverage to include more Southeast Asian languages, such as Malay, Lao, and Khmer.
    \item \textbf{Task Type}: At present, SailCompass includes tasks for text generation, multi-question answering, and classification. Future iterations should also encompass advanced tasks such as mathematics, coding, and other specialized domains.
\end{enumerate}

For the evaluation methods:
\begin{enumerate}
    \item \textbf{Prompt Construction}: SailCompass currently explores different prompt configurations, but the prompt templates are still manually constructed. Future work should focus on optimizing prompts for each model.
    \item \textbf{Calibration Methods}: SailCompass currently utilizes contextual calibration, which is more effective for one-token label predictions. Future research should explore Domain Conditional PMI, which is more general by removing surface form competition and addressing general output bias.
\end{enumerate}

\section*{Ethics Statement}
This work presents the SailCompass benchmark, which is built upon existing datasets. The related licenses of these datasets are all open for academic usage, ensuring compliance with their terms and conditions. We did not build any new datasets for this work.
All code, models, and data used in this research are publicly accessible. We are committed to open science and have made all our resources available to the community to facilitate further research and development in this field.

\bibliographystyle{plainnat}
\bibliography{references}

\begin{thebibliography}{48}
\providecommand{\natexlab}[1]{#1}
\providecommand{\url}[1]{\texttt{#1}}
\expandafter\ifx\csname urlstyle\endcsname\relax
  \providecommand{\doi}[1]{doi: #1}\else
  \providecommand{\doi}{doi: \begingroup \urlstyle{rm}\Url}\fi

\bibitem[Ahuja et~al.(2023)Ahuja, Diddee, Hada, Ochieng, Ramesh, Jain, Nambi, Ganu, Segal, Ahmed, Bali, and Sitaram]{MEGA23}
Kabir Ahuja, Harshita Diddee, Rishav Hada, Millicent Ochieng, Krithika Ramesh, Prachi Jain, Akshay~Uttama Nambi, Tanuja Ganu, Sameer Segal, Mohamed Ahmed, Kalika Bali, and Sunayana Sitaram.
\newblock {MEGA:} multilingual evaluation of generative {AI}.
\newblock In \emph{Proceedings of the 2023 Conference on Empirical Methods in Natural Language Processing, {EMNLP} 2023}, pages 4232--4267, 2023.
\newblock URL \url{https://doi.org/10.18653/v1/2023.emnlp-main.258}.

\bibitem[{AI Singapore}(2023)]{sealion2023}
{AI Singapore}.
\newblock Sea-lion (southeast asian languages in one network): A family of large language models for southeast asia.
\newblock \url{https://github.com/aisingapore/sealion}, 2023.

\bibitem[Aji et~al.(2022)Aji, Winata, Koto, Cahyawijaya, Romadhony, Mahendra, Kurniawan, Moeljadi, Prasojo, Baldwin, Lau, and Ruder]{aji-etal-2022-one}
Alham~Fikri Aji, Genta~Indra Winata, Fajri Koto, Samuel Cahyawijaya, Ade Romadhony, Rahmad Mahendra, Kemal Kurniawan, David Moeljadi, Radityo~Eko Prasojo, Timothy Baldwin, Jey~Han Lau, and Sebastian Ruder.
\newblock One country, 700+ languages: {NLP} challenges for underrepresented languages and dialects in {I}ndonesia.
\newblock In Smaranda Muresan, Preslav Nakov, and Aline Villavicencio, editors, \emph{Proceedings of the 60th Annual Meeting of the Association for Computational Linguistics (Volume 1: Long Papers)}, pages 7226--7249, Dublin, Ireland, May 2022. Association for Computational Linguistics.
\newblock \doi{10.18653/v1/2022.acl-long.500}.
\newblock URL \url{https://aclanthology.org/2022.acl-long.500}.

\bibitem[Artetxe et~al.(2020)Artetxe, Ruder, and Yogatama]{xquad2020}
Mikel Artetxe, Sebastian Ruder, and Dani Yogatama.
\newblock On the cross-lingual transferability of monolingual representations.
\newblock In \emph{Proceedings of the 58th Annual Meeting of the Association for Computational Linguistics, {ACL} 2020, Online, July 5-10, 2020}, pages 4623--4637. Association for Computational Linguistics, 2020.
\newblock URL \url{https://doi.org/10.18653/v1/2020.acl-main.421}.

\bibitem[Asai et~al.(2024)Asai, Kudugunta, Yu, Blevins, Gonen, Reid, Tsvetkov, Ruder, and Hajishirzi]{BUFFET24}
Akari Asai, Sneha Kudugunta, Xinyan~Velocity Yu, Terra Blevins, Hila~B Gonen, Machel Reid, Yulia Tsvetkov, Sebastian Ruder, and Hannaneh Hajishirzi.
\newblock {BUFFET}: Benchmarking large language models for cross-lingual few-shot transfer.
\newblock In \emph{NAACL}, 2024.

\bibitem[Bai et~al.(2023)Bai, Bai, Chu, Cui, Dang, Deng, Fan, Ge, Han, Huang, Hui, Ji, Li, Lin, Lin, Liu, Liu, Lu, Lu, Ma, Men, Ren, Ren, Tan, Tan, Tu, Wang, Wang, Wang, Wu, Xu, Xu, Yang, Yang, Yang, Yang, Yao, Yu, Yuan, Yuan, Zhang, Zhang, Zhang, Zhang, Zhou, Zhou, Zhou, and Zhu]{bai2023qwen}
Jinze Bai, Shuai Bai, Yunfei Chu, Zeyu Cui, Kai Dang, Xiaodong Deng, Yang Fan, Wenbin Ge, Yu~Han, Fei Huang, Binyuan Hui, Luo Ji, Mei Li, Junyang Lin, Runji Lin, Dayiheng Liu, Gao Liu, Chengqiang Lu, Keming Lu, Jianxin Ma, Rui Men, Xingzhang Ren, Xuancheng Ren, Chuanqi Tan, Sinan Tan, Jianhong Tu, Peng Wang, Shijie Wang, Wei Wang, Shengguang Wu, Benfeng Xu, Jin Xu, An~Yang, Hao Yang, Jian Yang, Shusheng Yang, Yang Yao, Bowen Yu, Hongyi Yuan, Zheng Yuan, Jianwei Zhang, Xingxuan Zhang, Yichang Zhang, Zhenru Zhang, Chang Zhou, Jingren Zhou, Xiaohuan Zhou, and Tianhang Zhu.
\newblock Qwen technical report, 2023.

\bibitem[Bandarkar et~al.(2023)Bandarkar, Liang, Muller, Artetxe, Shukla, Husa, Goyal, Krishnan, Zettlemoyer, and Khabsa]{belebele2023}
Lucas Bandarkar, Davis Liang, Benjamin Muller, Mikel Artetxe, Satya~Narayan Shukla, Donald Husa, Naman Goyal, Abhinandan Krishnan, Luke Zettlemoyer, and Madian Khabsa.
\newblock The belebele benchmark: a parallel reading comprehension dataset in 122 language variants.
\newblock \emph{CoRR}, abs/2308.16884, 2023.
\newblock URL \url{https://doi.org/10.48550/arXiv.2308.16884}.

\bibitem[Cahyawijaya et~al.(2023)Cahyawijaya, Lovenia, Aji, Winata, Wilie, Koto, Mahendra, Wibisono, Romadhony, Vincentio, Santoso, Moeljadi, Wirawan, Hudi, Wicaksono, Parmonangan, Alfina, Putra, Rahmadani, Oenang, Septiandri, Jaya, Dhole, Suryani, Putri, Su, Stevens, Nityasya, Adilazuarda, Hadiwijaya, Diandaru, Yu, Ghifari, Dai, Xu, Damapuspita, Wibowo, Tho, Karo, Fatyanosa, Ji, Neubig, Baldwin, Ruder, Fung, Sujaini, Sakti, and Purwarianti]{NusaCrowd2023}
Samuel Cahyawijaya, Holy Lovenia, Alham~Fikri Aji, Genta~Indra Winata, Bryan Wilie, Fajri Koto, Rahmad Mahendra, Christian Wibisono, Ade Romadhony, Karissa Vincentio, Jennifer Santoso, David Moeljadi, Cahya Wirawan, Frederikus Hudi, Muhammad~Satrio Wicaksono, Ivan~Halim Parmonangan, Ika Alfina, Ilham~Firdausi Putra, Samsul Rahmadani, Yulianti Oenang, Ali~Akbar Septiandri, James Jaya, Kaustubh~D. Dhole, Arie~Ardiyanti Suryani, Rifki~Afina Putri, Dan Su, Keith Stevens, Made~Nindyatama Nityasya, Muhammad~Farid Adilazuarda, Ryan Hadiwijaya, Ryandito Diandaru, Tiezheng Yu, Vito Ghifari, Wenliang Dai, Yan Xu, Dyah Damapuspita, Haryo~Akbarianto Wibowo, Cuk Tho, Ichwanul Muslim~Karo Karo, Tirana Fatyanosa, Ziwei Ji, Graham Neubig, Timothy Baldwin, Sebastian Ruder, Pascale Fung, Herry Sujaini, Sakriani Sakti, and Ayu Purwarianti.
\newblock Nusacrowd: Open source initiative for indonesian {NLP} resources.
\newblock In \emph{Findings of the Association for Computational Linguistics: {ACL} 2023, Toronto, Canada, July 9-14, 2023}, pages 13745--13818. Association for Computational Linguistics, 2023.
\newblock URL \url{https://doi.org/10.18653/v1/2023.findings-acl.868}.

\bibitem[Chen et~al.(2022)Chen, Song, Wu, Wang, Xu, Chen, Zhou, and Li]{MTG22}
Yiran Chen, Zhenqiao Song, Xianze Wu, Danqing Wang, Jingjing Xu, Jiaze Chen, Hao Zhou, and Lei Li.
\newblock {MTG:} {A} benchmark suite for multilingual text generation.
\newblock In \emph{Findings of the Association for Computational Linguistics: {NAACL} 2022}, pages 2508--2527, 2022.
\newblock URL \url{https://doi.org/10.18653/v1/2022.findings-naacl.192}.

\bibitem[Chumpolsathien(2020)]{thaisum2020}
Nakhun Chumpolsathien.
\newblock Using knowledge distillation from keyword extraction to improve the informativeness of neural cross-lingual summarization.
\newblock Master's thesis, Beijing Institute of Technology, 2020.

\bibitem[Clark et~al.(2020)Clark, Palomaki, Nikolaev, Choi, Garrette, Collins, and Kwiatkowski]{tydiqa2020}
Jonathan~H. Clark, Jennimaria Palomaki, Vitaly Nikolaev, Eunsol Choi, Dan Garrette, Michael Collins, and Tom Kwiatkowski.
\newblock Tydi {QA:} {A} benchmark for information-seeking question answering in typologically diverse languages.
\newblock \emph{Trans. Assoc. Comput. Linguistics}, 8:\penalty0 454--470, 2020.
\newblock URL \url{https://doi.org/10.1162/tacl\_a\_00317}.

\bibitem[Conneau et~al.(2018)Conneau, Rinott, Lample, Williams, Bowman, Schwenk, and Stoyanov]{xnli2018}
Alexis Conneau, Ruty Rinott, Guillaume Lample, Adina Williams, Samuel~R. Bowman, Holger Schwenk, and Veselin Stoyanov.
\newblock {XNLI:} evaluating cross-lingual sentence representations.
\newblock In \emph{Proceedings of the 2018 Conference on Empirical Methods in Natural Language Processing, Brussels, Belgium, October 31 - November 4, 2018}, pages 2475--2485. Association for Computational Linguistics, 2018.
\newblock URL \url{https://doi.org/10.18653/v1/d18-1269}.

\bibitem[Costa{-}juss{\`{a}} et~al.(2022)Costa{-}juss{\`{a}}, Cross, {\c{C}}elebi, Elbayad, Heafield, Heffernan, Kalbassi, Lam, Licht, Maillard, Sun, Wang, Wenzek, Youngblood, Akula, Barrault, Gonzalez, Hansanti, Hoffman, Jarrett, Sadagopan, Rowe, Spruit, Tran, Andrews, Ayan, Bhosale, Edunov, Fan, Gao, Goswami, Guzm{\'{a}}n, Koehn, Mourachko, Ropers, Saleem, Schwenk, and Wang]{flores2022}
Marta~R. Costa{-}juss{\`{a}}, James Cross, Onur {\c{C}}elebi, Maha Elbayad, Kenneth Heafield, Kevin Heffernan, Elahe Kalbassi, Janice Lam, Daniel Licht, Jean Maillard, Anna~Y. Sun, Skyler Wang, Guillaume Wenzek, Al~Youngblood, Bapi Akula, Lo{\"{\i}}c Barrault, Gabriel~Mejia Gonzalez, Prangthip Hansanti, John Hoffman, Semarley Jarrett, Kaushik~Ram Sadagopan, Dirk Rowe, Shannon Spruit, Chau Tran, Pierre Andrews, Necip~Fazil Ayan, Shruti Bhosale, Sergey Edunov, Angela Fan, Cynthia Gao, Vedanuj Goswami, Francisco Guzm{\'{a}}n, Philipp Koehn, Alexandre Mourachko, Christophe Ropers, Safiyyah Saleem, Holger Schwenk, and Jeff Wang.
\newblock No language left behind: Scaling human-centered machine translation.
\newblock \emph{CoRR}, abs/2207.04672, 2022.
\newblock URL \url{https://doi.org/10.48550/arXiv.2207.04672}.

\bibitem[Dou et~al.(2024)Dou, Liu, Zeng, Guo, Zhou, Lu, and Lin]{dou2024sailor}
Longxu Dou, Qian Liu, Guangtao Zeng, Jia Guo, Jiahui Zhou, Wei Lu, and Min Lin.
\newblock Sailor: Open language models for south-east asia.
\newblock \emph{arXiv preprint arXiv:2404.03608}, 2024.

\bibitem[Hasan et~al.(2021)Hasan, Bhattacharjee, Islam, Mubasshir, Li, Kang, Rahman, and Shahriyar]{xlsum21}
Tahmid Hasan, Abhik Bhattacharjee, Md.~Saiful Islam, Kazi Mubasshir, Yuan-Fang Li, Yong-Bin Kang, M.~Sohel Rahman, and Rifat Shahriyar.
\newblock {XL}-sum: Large-scale multilingual abstractive summarization for 44 languages.
\newblock In \emph{Findings of the Association for Computational Linguistics: ACL-IJCNLP 2021}, 2021.
\newblock URL \url{https://aclanthology.org/2021.findings-acl.413}.

\bibitem[Hendrycks et~al.(2021)Hendrycks, Burns, Basart, Zou, Mazeika, Song, and Steinhardt]{mmlu2021}
Dan Hendrycks, Collin Burns, Steven Basart, Andy Zou, Mantas Mazeika, Dawn Song, and Jacob Steinhardt.
\newblock Measuring massive multitask language understanding.
\newblock In \emph{9th International Conference on Learning Representations, {ICLR} 2021, Virtual Event, Austria, May 3-7, 2021}. OpenReview.net, 2021.
\newblock URL \url{https://openreview.net/forum?id=d7KBjmI3GmQ}.

\bibitem[Ho et~al.(2019)Ho, Nguyen, Nguyen, Pham, Nguyen, Nguyen, and Nguyen]{vsmec}
Vong~Anh Ho, Duong~Huynh{-}Cong Nguyen, Danh~Hoang Nguyen, Linh~Thi{-}Van Pham, Duc{-}Vu Nguyen, Kiet~Van Nguyen, and Ngan~Luu{-}Thuy Nguyen.
\newblock Emotion recognition for vietnamese social media text.
\newblock In \emph{Computational Linguistics - 16th International Conference of the Pacific Association for Computational Linguistics, {PACLING} 2019, Hanoi, Vietnam, October 11-13, 2019, Revised Selected Papers}, volume 1215 of \emph{Communications in Computer and Information Science}, pages 319--333. Springer, 2019.
\newblock URL \url{https://doi.org/10.1007/978-981-15-6168-9\_27}.

\bibitem[Hu et~al.(2020)Hu, Ruder, Siddhant, Neubig, Firat, and Johnson]{XTREME20}
Junjie Hu, Sebastian Ruder, Aditya Siddhant, Graham Neubig, Orhan Firat, and Melvin Johnson.
\newblock {XTREME:} {A} massively multilingual multi-task benchmark for evaluating cross-lingual generalization.
\newblock In \emph{Proceedings of the 37th International Conference on Machine Learning, {ICML} 2020}, Proceedings of Machine Learning Research, 2020.
\newblock URL \url{https://proceedings.mlr.press/v119/hu20b/hu20b.pdf}.

\bibitem[Jiang et~al.(2023)Jiang, Sablayrolles, Mensch, Bamford, Chaplot, de~las Casas, Bressand, Lengyel, Lample, Saulnier, Lavaud, Lachaux, Stock, Scao, Lavril, Wang, Lacroix, and Sayed]{jiang2023mistral}
Albert~Q. Jiang, Alexandre Sablayrolles, Arthur Mensch, Chris Bamford, Devendra~Singh Chaplot, Diego de~las Casas, Florian Bressand, Gianna Lengyel, Guillaume Lample, Lucile Saulnier, Lélio~Renard Lavaud, Marie-Anne Lachaux, Pierre Stock, Teven~Le Scao, Thibaut Lavril, Thomas Wang, Timothée Lacroix, and William~El Sayed.
\newblock Mistral 7b, 2023.

\bibitem[Koto et~al.(2020)Koto, Rahimi, Lau, and Baldwin]{indolemsenti2020}
Fajri Koto, Afshin Rahimi, Jey~Han Lau, and Timothy Baldwin.
\newblock Indolem and indobert: {A} benchmark dataset and pre-trained language model for indonesian {NLP}.
\newblock In \emph{Proceedings of the 28th International Conference on Computational Linguistics, {COLING} 2020, Barcelona, Spain (Online), December 8-13, 2020}, pages 757--770. International Committee on Computational Linguistics, 2020.
\newblock URL \url{https://doi.org/10.18653/v1/2020.coling-main.66}.

\bibitem[Kurniawan and Louvan(2018)]{indosum2018}
Kemal Kurniawan and Samuel Louvan.
\newblock Indosum: {A} new benchmark dataset for indonesian text summarization.
\newblock In \emph{2018 International Conference on Asian Language Processing, {IALP} 2018, Bandung, Indonesia, November 15-17, 2018}, pages 215--220. {IEEE}, 2018.
\newblock URL \url{https://doi.org/10.1109/IALP.2018.8629109}.

\bibitem[Leong et~al.(2023)Leong, Ngui, Susanto, Rengarajan, Sarveswaran, and Tjhi]{BHASA2023}
Wei~Qi Leong, Jian~Gang Ngui, Yosephine Susanto, Hamsawardhini Rengarajan, Kengatharaiyer Sarveswaran, and William{-}Chandra Tjhi.
\newblock {BHASA:} {A} holistic southeast asian linguistic and cultural evaluation suite for large language models.
\newblock \emph{CoRR}, abs/2309.06085, 2023.
\newblock URL \url{https://doi.org/10.48550/arXiv.2309.06085}.

\bibitem[Liang et~al.(2020)Liang, Duan, Gong, Wu, Guo, Qi, Gong, Shou, Jiang, Cao, Fan, Zhang, Agrawal, Cui, Wei, Bharti, Qiao, Chen, Wu, Liu, Yang, Campos, Majumder, and Zhou]{XGLUE20}
Yaobo Liang, Nan Duan, Yeyun Gong, Ning Wu, Fenfei Guo, Weizhen Qi, Ming Gong, Linjun Shou, Daxin Jiang, Guihong Cao, Xiaodong Fan, Ruofei Zhang, Rahul Agrawal, Edward Cui, Sining Wei, Taroon Bharti, Ying Qiao, Jiun{-}Hung Chen, Winnie Wu, Shuguang Liu, Fan Yang, Daniel Campos, Rangan Majumder, and Ming Zhou.
\newblock {XGLUE:} {A} new benchmark datasetfor cross-lingual pre-training, understanding and generation.
\newblock In \emph{Proceedings of the 2020 Conference on Empirical Methods in Natural Language Processing, {EMNLP} 2020}, pages 6008--6018, 2020.
\newblock URL \url{https://doi.org/10.18653/v1/2020.emnlp-main.484}.

\bibitem[Lu et~al.(2023)Lu, Huang, Zhang, Yang, Lam, and Wei]{Lu2023ChainofDictionaryPE}
Hongyuan Lu, Haoyang Huang, Dongdong Zhang, Haoran Yang, Wai Lam, and Furu Wei.
\newblock Chain-of-dictionary prompting elicits translation in large language models.
\newblock \emph{ArXiv}, abs/2305.06575, 2023.

\bibitem[Mahendra et~al.(2021)Mahendra, Aji, Louvan, Rahman, and Vania]{indonli2021}
Rahmad Mahendra, Alham~Fikri Aji, Samuel Louvan, Fahrurrozi Rahman, and Clara Vania.
\newblock Indonli: {A} natural language inference dataset for indonesian.
\newblock In \emph{Proceedings of the 2021 Conference on Empirical Methods in Natural Language Processing, {EMNLP} 2021, Virtual Event / Punta Cana, Dominican Republic, 7-11 November, 2021}, pages 10511--10527. Association for Computational Linguistics, 2021.
\newblock URL \url{https://doi.org/10.18653/v1/2021.emnlp-main.821}.

\bibitem[Nguyen et~al.(2023{\natexlab{a}})Nguyen, Pham, and Dao]{nguyen2023vinallama}
Quan Nguyen, Huy Pham, and Dung Dao.
\newblock Vinallama: Llama-based vietnamese foundation model, 2023{\natexlab{a}}.

\bibitem[Nguyen et~al.(2023{\natexlab{b}})Nguyen, Zhang, Li, Aljunied, Tan, Cheng, Chen, Deng, Yang, Liu, Zhang, and Bing]{seallm2023}
Xuan{-}Phi Nguyen, Wenxuan Zhang, Xin Li, Mahani Aljunied, Qingyu Tan, Liying Cheng, Guanzheng Chen, Yue Deng, Sen Yang, Chaoqun Liu, Hang Zhang, and Lidong Bing.
\newblock Seallms - large language models for southeast asia.
\newblock \emph{CoRR}, abs/2312.00738, 2023{\natexlab{b}}.
\newblock URL \url{https://doi.org/10.48550/arXiv.2312.00738}.

\bibitem[{OpenCompass Contributors}(2023)]{2023opencompass}
{OpenCompass Contributors}.
\newblock Opencompass: A universal evaluation platform for foundation models.
\newblock \url{https://github.com/open-compass/opencompass}, 2023.

\bibitem[Papineni et~al.(2002)Papineni, Roukos, Ward, and Zhu]{papineni-etal-2002-bleu}
Kishore Papineni, Salim Roukos, Todd Ward, and Wei-Jing Zhu.
\newblock {B}leu: a method for automatic evaluation of machine translation.
\newblock In Pierre Isabelle, Eugene Charniak, and Dekang Lin, editors, \emph{Proceedings of the 40th Annual Meeting of the Association for Computational Linguistics}, pages 311--318, Philadelphia, Pennsylvania, USA, July 2002. Association for Computational Linguistics.
\newblock \doi{10.3115/1073083.1073135}.
\newblock URL \url{https://aclanthology.org/P02-1040}.

\bibitem[Pipatanakul et~al.(2023)Pipatanakul, Jirabovonvisut, Manakul, Sripaisarnmongkol, Patomwong, Chokchainant, and Tharnpipitchai]{pipatanakul2023typhoon}
Kunat Pipatanakul, Phatrasek Jirabovonvisut, Potsawee Manakul, Sittipong Sripaisarnmongkol, Ruangsak Patomwong, Pathomporn Chokchainant, and Kasima Tharnpipitchai.
\newblock Typhoon: Thai large language models, 2023.

\bibitem[Ponti et~al.(2020)Ponti, Glavas, Majewska, Liu, Vulic, and Korhonen]{xcopa2020}
Edoardo~Maria Ponti, Goran Glavas, Olga Majewska, Qianchu Liu, Ivan Vulic, and Anna Korhonen.
\newblock {XCOPA:} {A} multilingual dataset for causal commonsense reasoning.
\newblock In \emph{Proceedings of the 2020 Conference on Empirical Methods in Natural Language Processing, {EMNLP} 2020, Online, November 16-20, 2020}, pages 2362--2376. Association for Computational Linguistics, 2020.
\newblock URL \url{https://doi.org/10.18653/v1/2020.emnlp-main.185}.

\bibitem[Popovi{\'c}(2015)]{popovic-2015-chrf}
Maja Popovi{\'c}.
\newblock chr{F}: character n-gram {F}-score for automatic {MT} evaluation.
\newblock In Ond{\v{r}}ej Bojar, Rajan Chatterjee, Christian Federmann, Barry Haddow, Chris Hokamp, Matthias Huck, Varvara Logacheva, and Pavel Pecina, editors, \emph{Proceedings of the Tenth Workshop on Statistical Machine Translation}, pages 392--395, Lisbon, Portugal, September 2015. Association for Computational Linguistics.
\newblock \doi{10.18653/v1/W15-3049}.
\newblock URL \url{https://aclanthology.org/W15-3049}.

\bibitem[Rajpurkar et~al.(2016)Rajpurkar, Zhang, Lopyrev, and Liang]{squad16}
Pranav Rajpurkar, Jian Zhang, Konstantin Lopyrev, and Percy Liang.
\newblock {SQ}u{AD}: 100,000+ questions for machine comprehension of text.
\newblock In \emph{Proceedings of the 2016 Conference on Empirical Methods in Natural Language Processing}, 2016.
\newblock URL \url{https://aclanthology.org/D16-1264}.

\bibitem[Robinson and Wingate(2023)]{MCP2023}
Joshua Robinson and David Wingate.
\newblock Leveraging large language models for multiple choice question answering.
\newblock In \emph{The Eleventh International Conference on Learning Representations, {ICLR} 2023, Kigali, Rwanda, May 1-5, 2023}. OpenReview.net, 2023.
\newblock URL \url{https://openreview.net/pdf?id=yKbprarjc5B}.

\bibitem[Sclar et~al.(2024)Sclar, Choi, Tsvetkov, and Suhr]{promptformatting2024}
Melanie Sclar, Yejin Choi, Yulia Tsvetkov, and Alane Suhr.
\newblock Quantifying language models' sensitivity to spurious features in prompt design or: How i learned to start worrying about prompt formatting.
\newblock In \emph{The Twelfth International Conference on Learning Representations}, 2024.
\newblock URL \url{https://openreview.net/forum?id=RIu5lyNXjT}.

\bibitem[Shi et~al.(2023)Shi, Chen, Misra, Scales, Dohan, Chi, Sch{\"{a}}rli, and Zhou]{irrelevantContext2023}
Freda Shi, Xinyun Chen, Kanishka Misra, Nathan Scales, David Dohan, Ed~H. Chi, Nathanael Sch{\"{a}}rli, and Denny Zhou.
\newblock Large language models can be easily distracted by irrelevant context.
\newblock In \emph{International Conference on Machine Learning, {ICML} 2023, 23-29 July 2023, Honolulu, Hawaii, {USA}}, volume 202 of \emph{Proceedings of Machine Learning Research}, pages 31210--31227. {PMLR}, 2023.
\newblock URL \url{https://proceedings.mlr.press/v202/shi23a.html}.

\bibitem[Suriyawongkul et~al.(2019)Suriyawongkul, Chuangsuwanich, Chormai, and Polpanumas]{wisesight2019}
Arthit Suriyawongkul, Ekapol Chuangsuwanich, Pattarawat Chormai, and Charin Polpanumas.
\newblock Pythainlp/wisesight-sentiment: First release, September 2019.
\newblock URL \url{https://doi.org/10.5281/zenodo.3457447}.

\bibitem[Team et~al.(2024)Team, Mesnard, Hardin, Dadashi, Bhupatiraju, Pathak, Sifre, Rivière, Kale, Love, Tafti, Hussenot, Sessa, Chowdhery, Roberts, Barua, Botev, Castro-Ros, Slone, Héliou, Tacchetti, Bulanova, Paterson, Tsai, Shahriari, Lan, Choquette-Choo, Crepy, Cer, Ippolito, Reid, Buchatskaya, Ni, Noland, Yan, Tucker, Muraru, Rozhdestvenskiy, Michalewski, Tenney, Grishchenko, Austin, Keeling, Labanowski, Lespiau, Stanway, Brennan, Chen, Ferret, Chiu, Mao-Jones, Lee, Yu, Millican, Sjoesund, Lee, Dixon, Reid, Mikuła, Wirth, Sharman, Chinaev, Thain, Bachem, Chang, Wahltinez, Bailey, Michel, Yotov, Chaabouni, Comanescu, Jana, Anil, McIlroy, Liu, Mullins, Smith, Borgeaud, Girgin, Douglas, Pandya, Shakeri, De, Klimenko, Hennigan, Feinberg, Stokowiec, hui Chen, Ahmed, Gong, Warkentin, Peran, Giang, Farabet, Vinyals, Dean, Kavukcuoglu, Hassabis, Ghahramani, Eck, Barral, Pereira, Collins, Joulin, Fiedel, Senter, Andreev, and Kenealy]{gemmateam2024gemma}
Gemma Team, Thomas Mesnard, Cassidy Hardin, Robert Dadashi, Surya Bhupatiraju, Shreya Pathak, Laurent Sifre, Morgane Rivière, Mihir~Sanjay Kale, Juliette Love, Pouya Tafti, Léonard Hussenot, Pier~Giuseppe Sessa, Aakanksha Chowdhery, Adam Roberts, Aditya Barua, Alex Botev, Alex Castro-Ros, Ambrose Slone, Amélie Héliou, Andrea Tacchetti, Anna Bulanova, Antonia Paterson, Beth Tsai, Bobak Shahriari, Charline~Le Lan, Christopher~A. Choquette-Choo, Clément Crepy, Daniel Cer, Daphne Ippolito, David Reid, Elena Buchatskaya, Eric Ni, Eric Noland, Geng Yan, George Tucker, George-Christian Muraru, Grigory Rozhdestvenskiy, Henryk Michalewski, Ian Tenney, Ivan Grishchenko, Jacob Austin, James Keeling, Jane Labanowski, Jean-Baptiste Lespiau, Jeff Stanway, Jenny Brennan, Jeremy Chen, Johan Ferret, Justin Chiu, Justin Mao-Jones, Katherine Lee, Kathy Yu, Katie Millican, Lars~Lowe Sjoesund, Lisa Lee, Lucas Dixon, Machel Reid, Maciej Mikuła, Mateo Wirth, Michael Sharman, Nikolai Chinaev, Nithum Thain, Olivier Bachem,
  Oscar Chang, Oscar Wahltinez, Paige Bailey, Paul Michel, Petko Yotov, Rahma Chaabouni, Ramona Comanescu, Reena Jana, Rohan Anil, Ross McIlroy, Ruibo Liu, Ryan Mullins, Samuel~L Smith, Sebastian Borgeaud, Sertan Girgin, Sholto Douglas, Shree Pandya, Siamak Shakeri, Soham De, Ted Klimenko, Tom Hennigan, Vlad Feinberg, Wojciech Stokowiec, Yu~hui Chen, Zafarali Ahmed, Zhitao Gong, Tris Warkentin, Ludovic Peran, Minh Giang, Clément Farabet, Oriol Vinyals, Jeff Dean, Koray Kavukcuoglu, Demis Hassabis, Zoubin Ghahramani, Douglas Eck, Joelle Barral, Fernando Pereira, Eli Collins, Armand Joulin, Noah Fiedel, Evan Senter, Alek Andreev, and Kathleen Kenealy.
\newblock Gemma: Open models based on gemini research and technology, 2024.

\bibitem[team et~al.(2022)team, Costa-juss{\`a}, Cross, cCelebi, Elbayad, Heafield, Heffernan, Kalbassi, Lam, Licht, Maillard, Sun, Wang, Wenzek, Youngblood, Akula, Barrault, Gonzalez, Hansanti, Hoffman, Jarrett, Sadagopan, Rowe, Spruit, Tran, Andrews, Ayan, Bhosale, Edunov, Fan, Gao, Goswami, Guzm'an, Koehn, Mourachko, Ropers, Saleem, Schwenk, and Wang]{team2022NoLL}
Nllb team, Marta~Ruiz Costa-juss{\`a}, James Cross, Onur cCelebi, Maha Elbayad, Kenneth Heafield, Kevin Heffernan, Elahe Kalbassi, Janice Lam, Daniel Licht, Jean Maillard, Anna Sun, Skyler Wang, Guillaume Wenzek, Alison Youngblood, Bapi Akula, Lo{\"i}c Barrault, Gabriel~Mejia Gonzalez, Prangthip Hansanti, John Hoffman, Semarley Jarrett, Kaushik~Ram Sadagopan, Dirk Rowe, Shannon~L. Spruit, C.~Tran, Pierre~Yves Andrews, Necip~Fazil Ayan, Shruti Bhosale, Sergey Edunov, Angela Fan, Cynthia Gao, Vedanuj Goswami, Francisco Guzm'an, Philipp Koehn, Alexandre Mourachko, Christophe Ropers, Safiyyah Saleem, Holger Schwenk, and Jeff Wang.
\newblock No language left behind: Scaling human-centered machine translation.
\newblock \emph{ArXiv}, abs/2207.04672, 2022.

\bibitem[Touvron et~al.(2023)Touvron, Martin, Stone, Albert, Almahairi, Babaei, Bashlykov, Batra, Bhargava, Bhosale, Bikel, Blecher, Ferrer, Chen, Cucurull, Esiobu, Fernandes, Fu, Fu, Fuller, Gao, Goswami, Goyal, Hartshorn, Hosseini, Hou, Inan, Kardas, Kerkez, Khabsa, Kloumann, Korenev, Koura, Lachaux, Lavril, Lee, Liskovich, Lu, Mao, Martinet, Mihaylov, Mishra, Molybog, Nie, Poulton, Reizenstein, Rungta, Saladi, Schelten, Silva, Smith, Subramanian, Tan, Tang, Taylor, Williams, Kuan, Xu, Yan, Zarov, Zhang, Fan, Kambadur, Narang, Rodriguez, Stojnic, Edunov, and Scialom]{touvron2023llama}
Hugo Touvron, Louis Martin, Kevin Stone, Peter Albert, Amjad Almahairi, Yasmine Babaei, Nikolay Bashlykov, Soumya Batra, Prajjwal Bhargava, Shruti Bhosale, Dan Bikel, Lukas Blecher, Cristian~Canton Ferrer, Moya Chen, Guillem Cucurull, David Esiobu, Jude Fernandes, Jeremy Fu, Wenyin Fu, Brian Fuller, Cynthia Gao, Vedanuj Goswami, Naman Goyal, Anthony Hartshorn, Saghar Hosseini, Rui Hou, Hakan Inan, Marcin Kardas, Viktor Kerkez, Madian Khabsa, Isabel Kloumann, Artem Korenev, Punit~Singh Koura, Marie-Anne Lachaux, Thibaut Lavril, Jenya Lee, Diana Liskovich, Yinghai Lu, Yuning Mao, Xavier Martinet, Todor Mihaylov, Pushkar Mishra, Igor Molybog, Yixin Nie, Andrew Poulton, Jeremy Reizenstein, Rashi Rungta, Kalyan Saladi, Alan Schelten, Ruan Silva, Eric~Michael Smith, Ranjan Subramanian, Xiaoqing~Ellen Tan, Binh Tang, Ross Taylor, Adina Williams, Jian~Xiang Kuan, Puxin Xu, Zheng Yan, Iliyan Zarov, Yuchen Zhang, Angela Fan, Melanie Kambadur, Sharan Narang, Aurelien Rodriguez, Robert Stojnic, Sergey Edunov, and Thomas
  Scialom.
\newblock Llama 2: Open foundation and fine-tuned chat models, 2023.

\bibitem[Wang et~al.(2019{\natexlab{a}})Wang, Pruksachatkun, Nangia, Singh, Michael, Hill, Levy, and Bowman]{SuperGLUE19}
Alex Wang, Yada Pruksachatkun, Nikita Nangia, Amanpreet Singh, Julian Michael, Felix Hill, Omer Levy, and Samuel~R. Bowman.
\newblock Superglue: {A} stickier benchmark for general-purpose language understanding systems.
\newblock In \emph{Advances in Neural Information Processing Systems 32: Annual Conference on Neural Information Processing Systems 2019, NeurIPS 2019}, pages 3261--3275, 2019{\natexlab{a}}.
\newblock URL \url{https://proceedings.neurips.cc/paper/2019/hash/4496bf24afe7fab6f046bf4923da8de6-Abstract.html}.

\bibitem[Wang et~al.(2019{\natexlab{b}})Wang, Singh, Michael, Hill, Levy, and Bowman]{GLUE19}
Alex Wang, Amanpreet Singh, Julian Michael, Felix Hill, Omer Levy, and Samuel~R. Bowman.
\newblock {GLUE:} {A} multi-task benchmark and analysis platform for natural language understanding.
\newblock In \emph{7th International Conference on Learning Representations, {ICLR} 2019}, 2019{\natexlab{b}}.
\newblock URL \url{https://openreview.net/forum?id=rJ4km2R5t7}.

\bibitem[Wang et~al.(2024{\natexlab{a}})Wang, Liu, Huang, Jiao, Ding, Aw, and Chen]{SeaEval2024}
Bin Wang, Zhengyuan Liu, Xin Huang, Fangkai Jiao, Yang Ding, Ai~Ti Aw, and Nancy~F. Chen.
\newblock Seaeval for multilingual foundation models: From cross-lingual alignment to cultural reasoning.
\newblock In \emph{NAACL}, 2024{\natexlab{a}}.

\bibitem[Wang et~al.(2024{\natexlab{b}})Wang, Ma, Hu, Weber-Genzel, Röttger, Kreuter, Hovy, and Plank]{firsttoken2024}
Xinpeng Wang, Bolei Ma, Chengzhi Hu, Leon Weber-Genzel, Paul Röttger, Frauke Kreuter, Dirk Hovy, and Barbara Plank.
\newblock "my answer is c": First-token probabilities do not match text answers in instruction-tuned language models, 2024{\natexlab{b}}.

\bibitem[Workshop et~al.(2023)Workshop, :, Scao, Fan, Akiki, Pavlick, Ilić, Hesslow, Castagné, Luccioni, Yvon, Gallé, Tow, Rush, Biderman, Webson, Ammanamanchi, Wang, Sagot, Muennighoff, del Moral, Ruwase, Bawden, Bekman, McMillan-Major, Beltagy, Nguyen, Saulnier, Tan, Suarez, Sanh, Laurençon, Jernite, Launay, Mitchell, Raffel, Gokaslan, Simhi, Soroa, Aji, Alfassy, Rogers, Nitzav, Xu, Mou, Emezue, Klamm, Leong, van Strien, Adelani, Radev, Ponferrada, Levkovizh, Kim, Natan, Toni, Dupont, Kruszewski, Pistilli, Elsahar, Benyamina, Tran, Yu, Abdulmumin, Johnson, Gonzalez-Dios, de~la Rosa, Chim, Dodge, Zhu, Chang, Frohberg, Tobing, Bhattacharjee, Almubarak, Chen, Lo, Werra, Weber, Phan, allal, Tanguy, Dey, Muñoz, Masoud, Grandury, Šaško, Huang, Coavoux, Singh, Jiang, Vu, Jauhar, Ghaleb, Subramani, Kassner, Khamis, Nguyen, Espejel, de~Gibert, Villegas, Henderson, Colombo, Amuok, Lhoest, Harliman, Bommasani, López, Ribeiro, Osei, Pyysalo, Nagel, Bose, Muhammad, Sharma, Longpre, Nikpoor, Silberberg, Pai,
  Zink, Torrent, Schick, Thrush, Danchev, Nikoulina, Laippala, Lepercq, Prabhu, Alyafeai, Talat, Raja, Heinzerling, Si, Taşar, Salesky, Mielke, Lee, Sharma, Santilli, Chaffin, Stiegler, Datta, Szczechla, Chhablani, Wang, Pandey, Strobelt, Fries, Rozen, Gao, Sutawika, Bari, Al-shaibani, Manica, Nayak, Teehan, Albanie, Shen, Ben-David, Bach, Kim, Bers, Fevry, Neeraj, Thakker, Raunak, Tang, Yong, Sun, Brody, Uri, Tojarieh, Roberts, Chung, Tae, Phang, Press, Li, Narayanan, Bourfoune, Casper, Rasley, Ryabinin, Mishra, Zhang, Shoeybi, Peyrounette, Patry, Tazi, Sanseviero, von Platen, Cornette, Lavallée, Lacroix, Rajbhandari, Gandhi, Smith, Requena, Patil, Dettmers, Baruwa, Singh, Cheveleva, Ligozat, Subramonian, Névéol, Lovering, Garrette, Tunuguntla, Reiter, Taktasheva, Voloshina, Bogdanov, Winata, Schoelkopf, Kalo, Novikova, Forde, Clive, Kasai, Kawamura, Hazan, Carpuat, Clinciu, Kim, Cheng, Serikov, Antverg, van~der Wal, Zhang, Zhang, Gehrmann, Mirkin, Pais, Shavrina, Scialom, Yun, Limisiewicz, Rieser,
  Protasov, Mikhailov, Pruksachatkun, Belinkov, Bamberger, Kasner, Rueda, Pestana, Feizpour, Khan, Faranak, Santos, Hevia, Unldreaj, Aghagol, Abdollahi, Tammour, HajiHosseini, Behroozi, Ajibade, Saxena, Ferrandis, McDuff, Contractor, Lansky, David, Kiela, Nguyen, Tan, Baylor, Ozoani, Mirza, Ononiwu, Rezanejad, Jones, Bhattacharya, Solaiman, Sedenko, Nejadgholi, Passmore, Seltzer, Sanz, Dutra, Samagaio, Elbadri, Mieskes, Gerchick, Akinlolu, McKenna, Qiu, Ghauri, Burynok, Abrar, Rajani, Elkott, Fahmy, Samuel, An, Kromann, Hao, Alizadeh, Shubber, Wang, Roy, Viguier, Le, Oyebade, Le, Yang, Nguyen, Kashyap, Palasciano, Callahan, Shukla, Miranda-Escalada, Singh, Beilharz, Wang, Brito, Zhou, Jain, Xu, Fourrier, Periñán, Molano, Yu, Manjavacas, Barth, Fuhrimann, Altay, Bayrak, Burns, Vrabec, Bello, Dash, Kang, Giorgi, Golde, Posada, Sivaraman, Bulchandani, Liu, Shinzato, de~Bykhovetz, Takeuchi, Pàmies, Castillo, Nezhurina, Sänger, Samwald, Cullan, Weinberg, Wolf, Mihaljcic, Liu, Freidank, Kang, Seelam, Dahlberg,
  Broad, Muellner, Fung, Haller, Chandrasekhar, Eisenberg, Martin, Canalli, Su, Su, Cahyawijaya, Garda, Deshmukh, Mishra, Kiblawi, Ott, Sang-aroonsiri, Kumar, Schweter, Bharati, Laud, Gigant, Kainuma, Kusa, Labrak, Bajaj, Venkatraman, Xu, Xu, Xu, Tan, Xie, Ye, Bras, Belkada, and Wolf]{workshop2023bloom}
BigScience Workshop, :, Teven~Le Scao, Angela Fan, Christopher Akiki, Ellie Pavlick, Suzana Ilić, Daniel Hesslow, Roman Castagné, Alexandra~Sasha Luccioni, François Yvon, Matthias Gallé, Jonathan Tow, Alexander~M. Rush, Stella Biderman, Albert Webson, Pawan~Sasanka Ammanamanchi, Thomas Wang, Benoît Sagot, Niklas Muennighoff, Albert~Villanova del Moral, Olatunji Ruwase, Rachel Bawden, Stas Bekman, Angelina McMillan-Major, Iz~Beltagy, Huu Nguyen, Lucile Saulnier, Samson Tan, Pedro~Ortiz Suarez, Victor Sanh, Hugo Laurençon, Yacine Jernite, Julien Launay, Margaret Mitchell, Colin Raffel, Aaron Gokaslan, Adi Simhi, Aitor Soroa, Alham~Fikri Aji, Amit Alfassy, Anna Rogers, Ariel~Kreisberg Nitzav, Canwen Xu, Chenghao Mou, Chris Emezue, Christopher Klamm, Colin Leong, Daniel van Strien, David~Ifeoluwa Adelani, Dragomir Radev, Eduardo~González Ponferrada, Efrat Levkovizh, Ethan Kim, Eyal~Bar Natan, Francesco~De Toni, Gérard Dupont, Germán Kruszewski, Giada Pistilli, Hady Elsahar, Hamza Benyamina, Hieu Tran,
  Ian Yu, Idris Abdulmumin, Isaac Johnson, Itziar Gonzalez-Dios, Javier de~la Rosa, Jenny Chim, Jesse Dodge, Jian Zhu, Jonathan Chang, Jörg Frohberg, Joseph Tobing, Joydeep Bhattacharjee, Khalid Almubarak, Kimbo Chen, Kyle Lo, Leandro~Von Werra, Leon Weber, Long Phan, Loubna~Ben allal, Ludovic Tanguy, Manan Dey, Manuel~Romero Muñoz, Maraim Masoud, María Grandury, Mario Šaško, Max Huang, Maximin Coavoux, Mayank Singh, Mike Tian-Jian Jiang, Minh~Chien Vu, Mohammad~A. Jauhar, Mustafa Ghaleb, Nishant Subramani, Nora Kassner, Nurulaqilla Khamis, Olivier Nguyen, Omar Espejel, Ona de~Gibert, Paulo Villegas, Peter Henderson, Pierre Colombo, Priscilla Amuok, Quentin Lhoest, Rheza Harliman, Rishi Bommasani, Roberto~Luis López, Rui Ribeiro, Salomey Osei, Sampo Pyysalo, Sebastian Nagel, Shamik Bose, Shamsuddeen~Hassan Muhammad, Shanya Sharma, Shayne Longpre, Somaieh Nikpoor, Stanislav Silberberg, Suhas Pai, Sydney Zink, Tiago~Timponi Torrent, Timo Schick, Tristan Thrush, Valentin Danchev, Vassilina Nikoulina,
  Veronika Laippala, Violette Lepercq, Vrinda Prabhu, Zaid Alyafeai, Zeerak Talat, Arun Raja, Benjamin Heinzerling, Chenglei Si, Davut~Emre Taşar, Elizabeth Salesky, Sabrina~J. Mielke, Wilson~Y. Lee, Abheesht Sharma, Andrea Santilli, Antoine Chaffin, Arnaud Stiegler, Debajyoti Datta, Eliza Szczechla, Gunjan Chhablani, Han Wang, Harshit Pandey, Hendrik Strobelt, Jason~Alan Fries, Jos Rozen, Leo Gao, Lintang Sutawika, M~Saiful Bari, Maged~S. Al-shaibani, Matteo Manica, Nihal Nayak, Ryan Teehan, Samuel Albanie, Sheng Shen, Srulik Ben-David, Stephen~H. Bach, Taewoon Kim, Tali Bers, Thibault Fevry, Trishala Neeraj, Urmish Thakker, Vikas Raunak, Xiangru Tang, Zheng-Xin Yong, Zhiqing Sun, Shaked Brody, Yallow Uri, Hadar Tojarieh, Adam Roberts, Hyung~Won Chung, Jaesung Tae, Jason Phang, Ofir Press, Conglong Li, Deepak Narayanan, Hatim Bourfoune, Jared Casper, Jeff Rasley, Max Ryabinin, Mayank Mishra, Minjia Zhang, Mohammad Shoeybi, Myriam Peyrounette, Nicolas Patry, Nouamane Tazi, Omar Sanseviero, Patrick von
  Platen, Pierre Cornette, Pierre~François Lavallée, Rémi Lacroix, Samyam Rajbhandari, Sanchit Gandhi, Shaden Smith, Stéphane Requena, Suraj Patil, Tim Dettmers, Ahmed Baruwa, Amanpreet Singh, Anastasia Cheveleva, Anne-Laure Ligozat, Arjun Subramonian, Aurélie Névéol, Charles Lovering, Dan Garrette, Deepak Tunuguntla, Ehud Reiter, Ekaterina Taktasheva, Ekaterina Voloshina, Eli Bogdanov, Genta~Indra Winata, Hailey Schoelkopf, Jan-Christoph Kalo, Jekaterina Novikova, Jessica~Zosa Forde, Jordan Clive, Jungo Kasai, Ken Kawamura, Liam Hazan, Marine Carpuat, Miruna Clinciu, Najoung Kim, Newton Cheng, Oleg Serikov, Omer Antverg, Oskar van~der Wal, Rui Zhang, Ruochen Zhang, Sebastian Gehrmann, Shachar Mirkin, Shani Pais, Tatiana Shavrina, Thomas Scialom, Tian Yun, Tomasz Limisiewicz, Verena Rieser, Vitaly Protasov, Vladislav Mikhailov, Yada Pruksachatkun, Yonatan Belinkov, Zachary Bamberger, Zdeněk Kasner, Alice Rueda, Amanda Pestana, Amir Feizpour, Ammar Khan, Amy Faranak, Ana Santos, Anthony Hevia, Antigona
  Unldreaj, Arash Aghagol, Arezoo Abdollahi, Aycha Tammour, Azadeh HajiHosseini, Bahareh Behroozi, Benjamin Ajibade, Bharat Saxena, Carlos~Muñoz Ferrandis, Daniel McDuff, Danish Contractor, David Lansky, Davis David, Douwe Kiela, Duong~A. Nguyen, Edward Tan, Emi Baylor, Ezinwanne Ozoani, Fatima Mirza, Frankline Ononiwu, Habib Rezanejad, Hessie Jones, Indrani Bhattacharya, Irene Solaiman, Irina Sedenko, Isar Nejadgholi, Jesse Passmore, Josh Seltzer, Julio~Bonis Sanz, Livia Dutra, Mairon Samagaio, Maraim Elbadri, Margot Mieskes, Marissa Gerchick, Martha Akinlolu, Michael McKenna, Mike Qiu, Muhammed Ghauri, Mykola Burynok, Nafis Abrar, Nazneen Rajani, Nour Elkott, Nour Fahmy, Olanrewaju Samuel, Ran An, Rasmus Kromann, Ryan Hao, Samira Alizadeh, Sarmad Shubber, Silas Wang, Sourav Roy, Sylvain Viguier, Thanh Le, Tobi Oyebade, Trieu Le, Yoyo Yang, Zach Nguyen, Abhinav~Ramesh Kashyap, Alfredo Palasciano, Alison Callahan, Anima Shukla, Antonio Miranda-Escalada, Ayush Singh, Benjamin Beilharz, Bo~Wang, Caio Brito,
  Chenxi Zhou, Chirag Jain, Chuxin Xu, Clémentine Fourrier, Daniel~León Periñán, Daniel Molano, Dian Yu, Enrique Manjavacas, Fabio Barth, Florian Fuhrimann, Gabriel Altay, Giyaseddin Bayrak, Gully Burns, Helena~U. Vrabec, Imane Bello, Ishani Dash, Jihyun Kang, John Giorgi, Jonas Golde, Jose~David Posada, Karthik~Rangasai Sivaraman, Lokesh Bulchandani, Lu~Liu, Luisa Shinzato, Madeleine~Hahn de~Bykhovetz, Maiko Takeuchi, Marc Pàmies, Maria~A Castillo, Marianna Nezhurina, Mario Sänger, Matthias Samwald, Michael Cullan, Michael Weinberg, Michiel~De Wolf, Mina Mihaljcic, Minna Liu, Moritz Freidank, Myungsun Kang, Natasha Seelam, Nathan Dahlberg, Nicholas~Michio Broad, Nikolaus Muellner, Pascale Fung, Patrick Haller, Ramya Chandrasekhar, Renata Eisenberg, Robert Martin, Rodrigo Canalli, Rosaline Su, Ruisi Su, Samuel Cahyawijaya, Samuele Garda, Shlok~S Deshmukh, Shubhanshu Mishra, Sid Kiblawi, Simon Ott, Sinee Sang-aroonsiri, Srishti Kumar, Stefan Schweter, Sushil Bharati, Tanmay Laud, Théo Gigant, Tomoya
  Kainuma, Wojciech Kusa, Yanis Labrak, Yash~Shailesh Bajaj, Yash Venkatraman, Yifan Xu, Yingxin Xu, Yu~Xu, Zhe Tan, Zhongli Xie, Zifan Ye, Mathilde Bras, Younes Belkada, and Thomas Wolf.
\newblock Bloom: A 176b-parameter open-access multilingual language model, 2023.

\bibitem[Zhang et~al.(2023)Zhang, Aljunied, Gao, Chia, and Bing]{m3exam2023}
Wenxuan Zhang, Mahani Aljunied, Chang Gao, Yew~Ken Chia, and Lidong Bing.
\newblock M3exam: {A} multilingual, multimodal, multilevel benchmark for examining large language models.
\newblock In \emph{Advances in Neural Information Processing Systems 36: Annual Conference on Neural Information Processing Systems 2023, NeurIPS 2023, New Orleans, LA, USA, December 10 - 16, 2023}, 2023.
\newblock URL \url{http://papers.nips.cc/paper\_files/paper/2023/hash/117c5c8622b0d539f74f6d1fb082a2e9-Abstract-Datasets\_and\_Benchmarks.html}.

\bibitem[Zhao et~al.(2021)Zhao, Wallace, Feng, Klein, and Singh]{Zhao2021CalibrateBU}
Tony Zhao, Eric Wallace, Shi Feng, Dan Klein, and Sameer Singh.
\newblock Calibrate before use: Improving few-shot performance of language models.
\newblock In \emph{International Conference on Machine Learning}, 2021.

\bibitem[Zheng et~al.(2024)Zheng, Zhou, Meng, Zhou, and Huang]{PriDe2024}
Chujie Zheng, Hao Zhou, Fandong Meng, Jie Zhou, and Minlie Huang.
\newblock Large language models are not robust multiple choice selectors.
\newblock In \emph{The Twelfth International Conference on Learning Representations}, 2024.
\newblock URL \url{https://openreview.net/forum?id=shr9PXz7T0}.

\end{thebibliography}

\newpage

\appendix
\section{Prompt Variants for MCQs}
\label{promp_variants}

\begin{figure}[ht]
\centering
    \scalebox{0.85}{
    \begin{tabular}{p{1.1\linewidth}}
      \toprule
      \texttt{\textcolor{blue}{Please follow the given examples, read the context, and answer the question.}} \\
      \\
      \texttt{\textcolor{blue}{[ in-context examples ]}} \\
      \\
      \texttt{\textcolor{blue}{Context: }Middle distance running is a relatively inexpensive sport; however, there are many misconceptions regarding the few pieces of equipment required to participate. Products can be purchased as needed, but most will have little or no real impact on performance. Athletes may feel that they prefer a product even when it provides no real benefits.} \\
      \texttt{\textcolor{blue}{Question: }According to the passage, why might a middle distance runner purchase a more expensive piece of equipment?} \\
      \textcolor{gray}{\texttt{A. It’s their personal preference}} \\
      \textcolor{gray}{\texttt{B. It has proven benefits}} \\
      \textcolor{gray}{\texttt{C. It will greatly impact their performance}} \\
      \textcolor{gray}{\texttt{D. There are misconceptions surrounding less expensive equipment}} \\
      \texttt{\textcolor{blue}{Answer:} \textcolor{gray}{A.} \textcolor{red}{It’s their personal preference}} \\
      \bottomrule
    \end{tabular}
    }
\caption{
The illustration of prompt configuration ${T_o}$. 
Note that the \textcolor{gray}{gray} text is NOT used in this configuration.
}
\label{sa}
\end{figure}

\begin{figure}[ht]
\centering
    \scalebox{0.85}{
    \begin{tabular}{p{1.1\linewidth}}
      \toprule
      \texttt{\textcolor{blue}{Please follow the given examples, read the context, and answer the question.}} \\
      \\
      \texttt{\textcolor{blue}{[ in-context examples ]}} \\
      \\
      \texttt{\textcolor{blue}{Context: }Middle distance running is a relatively inexpensive sport; however, there are many misconceptions regarding the few pieces of equipment required to participate. Products can be purchased as needed, but most will have little or no real impact on performance. Athletes may feel that they prefer a product even when it provides no real benefits.} \\
      \texttt{\textcolor{blue}{Question: }According to the passage, why might a middle distance runner purchase a more expensive piece of equipment?} \\
      \texttt{It’s their personal preference} \\
      \texttt{It has proven benefits} \\
      \texttt{It will greatly impact their performance} \\
      \texttt{There are misconceptions surrounding less expensive equipment} \\
      \texttt{\textcolor{blue}{Answer:} \textcolor{red}{It’s their personal preference}} \\
      \bottomrule
    \end{tabular}
    }
\caption{The illustration of prompt configuration ${T_iT_o}$.
}
\label{sosa}
\end{figure}

\begin{figure}[ht]
\centering
    \scalebox{0.85}{
    \begin{tabular}{p{1.1\linewidth}}
      \toprule
      \texttt{\textcolor{blue}{Please follow the given examples, read the context, and answer the question.}} \\
      \\
      \texttt{\textcolor{blue}{[ in-context examples ]}} \\
      \\
      \texttt{\textcolor{blue}{Context: }Middle distance running is a relatively inexpensive sport; however, there are many misconceptions regarding the few pieces of equipment required to participate. Products can be purchased as needed, but most will have little or no real impact on performance. Athletes may feel that they prefer a product even when it provides no real benefits.} \\
      \texttt{\textcolor{blue}{Question: }According to the passage, why might a middle distance runner purchase a more expensive piece of equipment?} \\
      \texttt{A. It’s their personal preference} \\
      \texttt{B. It has proven benefits} \\
      \texttt{C. It will greatly impact their performance} \\
      \texttt{D. There are misconceptions surrounding less expensive equipment} \\
      \texttt{\textcolor{blue}{Answer:} \textcolor{red}{It’s their personal preference}} \\
      \bottomrule
    \end{tabular}
    }
\caption{The illustration of prompt configuration ${L_iT_iT_o}$.}
\label{iososa}
\end{figure}

\begin{figure}[ht]
\centering
    \scalebox{0.85}{
    \begin{tabular}{p{1.1\linewidth}}
      \toprule
      \texttt{\textcolor{blue}{Please follow the given examples, read the context, and answer the question.}} \\
      \\
      \texttt{\textcolor{blue}{[ in-context examples ]}} \\
      \\
      \texttt{\textcolor{blue}{Context: }Middle distance running is a relatively inexpensive sport; however, there are many misconceptions regarding the few pieces of equipment required to participate. Products can be purchased as needed, but most will have little or no real impact on performance. Athletes may feel that they prefer a product even when it provides no real benefits.} \\
      \texttt{\textcolor{blue}{Question: }According to the passage, why might a middle distance runner purchase a more expensive piece of equipment?} \\
      \texttt{A. It’s their personal preference} \\
      \texttt{B. It has proven benefits} \\
      \texttt{C. It will greatly impact their performance} \\
      \texttt{D. There are misconceptions surrounding less expensive equipment} \\
      \texttt{\textcolor{blue}{Answer:} \textcolor{red}{A}} \\
      \bottomrule
    \end{tabular}
    }
\caption{The illustration of prompt configuration ${L_iT_iL_o}$.}
\label{iosoia}
\end{figure}

\begin{figure}[ht]
\centering
    \scalebox{0.85}{
    \begin{tabular}{p{1.1\linewidth}}
      \toprule
      \texttt{\textcolor{blue}{Please follow the given examples, read the context, and answer the question.}} \\
      \\
      \texttt{\textcolor{blue}{[ in-context examples ]}} \\
      \\
      \texttt{\textcolor{blue}{Context: }Middle distance running is a relatively inexpensive sport; however, there are many misconceptions regarding the few pieces of equipment required to participate. Products can be purchased as needed, but most will have little or no real impact on performance. Athletes may feel that they prefer a product even when it provides no real benefits.} \\
      \texttt{\textcolor{blue}{Question: }According to the passage, why might a middle distance runner purchase a more expensive piece of equipment?} \\
      \texttt{A. It’s their personal preference} \\
      \texttt{B. It has proven benefits} \\
      \texttt{C. It will greatly impact their performance} \\
      \texttt{D. There are misconceptions surrounding less expensive equipment} \\
      \texttt{\textcolor{blue}{Answer:} \textcolor{red}{A. It’s their personal preference}} \\
      \bottomrule
    \end{tabular}
    }
\caption{The illustration of prompt configuration ${L_iT_iL_oT_o}$.}
\label{iosoiasa}
\end{figure}

\section{Related Work on Multilingual Benchmarks}\label{appendix:multilingual_benchmarks}
Existing English multi-task benchmarks, such as GLUE~\cite{GLUE19} and SuperGLUE~\cite{SuperGLUE19}, have undoubtedly stimulated the growth in research interest and efforts on the transfer learning ability of language models across diverse tasks. However, the development of multilingual benchmarks has significantly lagged behind that of English-dominant benchmarks. 
To fill this gap, XTREME~\cite{XTREME20} contributed a comprehensive multilingual multi-task benchmark for evaluating cross-lingual transfer learning across 40 languages with 9 datasets. 
MTG~\cite{MTG22} included four human-annotated text generation datasets in five languages to support both training and test scenarios.

XGLUE~\cite{XGLUE20} expanded the task scope of the multilingual benchmark to encompass both natural language understanding and generation tasks. 
SeaEval~\cite{SeaEval2024} provided a benchmark for multilingual foundation models, additionally considering the cultural understanding ability of models, but the majority of the datasets were still based on high-resource languages, such as English and Chinese. 
The aforementioned benchmark either neglects tasks in Southeast Asian languages (e.g., MTG and XGLUE), or only contains a limited subset of tasks or languages from Southeast Asia (e.g., XTREME and SeaEval). 

Recent multilingual benchmarks like MEGA~\cite{MEGA23} and BUFFET~\cite{BUFFET24}, narrow the datasets that built from scratch in native languages and the models that specific designed to Southeast Asia (SEA). 
This limits our ability to draw comprehensive and systematic conclusions for SEA large language model research. 
In comparison, we broaden the evaluation scope by considering more models and expanding the datasets.

\section{Manipulated Training Details}\label{appendix:manipulate_exp} 
We adopt the ``auxiliary train'' split of the MMLU dataset \citep{mmlu2021} as the training corpus \footnote{https://huggingface.co/datasets/cais/mmlu/viewer/auxiliary\_train}, which contains 99.8k examples. 
After removing the duplicate examples, it resulted in 98.4k examples. 
We formulate the training dataset to align with the corresponding format of prompt variants. During training, we select the latest LLM for Southeast Asian languages for this experiment. The context window is set to 4096. The warmup step is 40. The cosine learning is scheduled with a maximum learning rate of 1e-5 and the weight decay is set to 0.1. We finetune the base model \verb|Sailor-7B| with 8 A100 GPUs. The total batch size is 512 and we train the model for 2 epochs with a total of 390 steps.

\begin{table}[t]
\small
\centering
\caption{
Model and their corresponding link used in our experiments.
}
\label{table:model_link}
\begin{tabular}{ll}
\toprule
\textbf{Model} & \textbf{Link} \\
\midrule
\texttt{Qwen-1.5-7B} &  \url{https://huggingface.co/Qwen/Qwen1.5-7B} \\
\texttt{Llama-2-7B} & \url{https://huggingface.co/meta-llama/Llama-2-7b-hf} \\
\texttt{Llama-3-8B} & \url{https://huggingface.co/meta-llama/Meta-Llama-3-8B} \\
\texttt{Mistral-7B} & \url{https://huggingface.co/mistralai/Mistral-7B-v0.1} \\
\texttt{Gemma-7B} & \url{https://huggingface.co/google/gemma-7b} \\
\texttt{Typhoon-8B} & \url{https://huggingface.co/scb10x/llama-3-typhoon-v1.5-8b}  \\
\texttt{VinaLLaMA-7B} & \url{https://huggingface.co/vilm/vinallama-7b}  \\
\texttt{BLOOM-7B1} & \url{https://huggingface.co/bigscience/bloom-7b1} \\
\texttt{Sailor-7B} & \url{https://huggingface.co/sail/Sailor-7B} \\
\texttt{SeaLLM-7B} & \url{https://huggingface.co/SeaLLMs/SeaLLM-7B-Hybrid}  \\
\texttt{Sea-Lion-7B} & \url{https://huggingface.co/aisingapore/sea-lion-7b}\\
\bottomrule
\end{tabular}
\end{table}

\section{Calibration Details}\label{appendix:calibration} 
In this section, we list the information of the label prediction before and after calibration, which are presented in Table~\ref{cls_task_label_th},~\ref{cls_task_label_id} and~\ref{cls_task_label_vi}.
We observe that models tend to predict only one or two labels, indicating a severe label prediction imbalance.
However, after applying Contextual Calibration methods, this imbalance can be mitigated a lot.

\begin{table}[t]
\begin{center}
\caption{
Thai NLI task results
}
\label{cls_task_label_th}
\scalebox{0.8}{
\begin{tabular}{crrr}
\toprule
& \textsc{contradiction} & \textsc{entailment}& \textsc{neutral}  \\
\midrule
\rowcolor{gray!30}
Gold &  1670	&1670	&1670 \\
\texttt{Seallm-7B} &  2752	& 2258 &	0 \\
\texttt{Sailor-7B} & 5007	& 3	 & 0 \\
\texttt{Sealion-7B} & 5010	& 0 &	0 \\
\midrule
\texttt{Seallm-7B}\_{\texttt{CC}} & 284	&3456& 	1270 \\
\texttt{Sailor-7B}\_{\texttt{CC}} & 3564	& 140	& 1306 \\
\texttt{Sealion-7B}\_{\texttt{CC}} & 3907& 933	& 170 \\
\bottomrule
\end{tabular}
}
\end{center}
\end{table}

\begin{table}[t]
\begin{center}
\caption{
Indonesian NLI task results.
}
\label{cls_task_label_id}
\scalebox{0.8}{
\begin{tabular}{crrr}
\toprule
& \textsc{contradiction} & \textsc{entailment}& \textsc{neutral}  \\
\midrule
\rowcolor{gray!30}
Gold &  1762&	1848	&1572 \\
\texttt{Seallm-7B} & 5152	&0	&30 \\
\texttt{Sailor-7B} & 5105	&77&	0 \\
\texttt{Sealion-7B} & 5182	& 0 &	0 \\
\midrule
\texttt{Seallm-7B}\_{\texttt{CC}} & 4060	&726	&396 \\
\texttt{Sailor-7B}\_{\texttt{CC}} & 4181&	396&	605 \\
\texttt{Sealion-7B}\_{\texttt{CC}} & 4678	&458&	46 \\
\bottomrule
\end{tabular}
}
\end{center}
\end{table}

\begin{table}[t]
\begin{center}
\caption{
Vietnamese NLI task results
}
\label{cls_task_label_vi}
\scalebox{0.8}{
\begin{tabular}{crrr}
\toprule
& \textsc{contradiction} & \textsc{entailment}& \textsc{neutral}  \\
\midrule
\rowcolor{gray!30}
Gold &  1670	&1670	&1670 \\
\texttt{Seallm-7B} &  5010	&0	&0 \\
\texttt{Sailor-7B} & 3711	&0	&1299 \\
\texttt{Sealion-7B} & 9	&0	&5001 \\
\midrule
\texttt{Seallm-7B}\_{\texttt{CC}} & 2974 &	2020&	16 \\
\texttt{Sailor-7B}\_{\texttt{CC}} & 3481&	1374	&155 \\
\texttt{Sealion-7B}\_{\texttt{CC}} & 4696	&31	&283 \\
\bottomrule
\end{tabular}
}
\end{center}
\end{table}

\end{document}